\documentclass[10pt,twocolumn,letterpaper]{article}

\usepackage{iccv}
\usepackage{times}
\usepackage{epsfig}
\usepackage{graphicx}
\usepackage{amsmath}
\usepackage{amssymb}
\usepackage{dsfont}
\usepackage{enumitem}
\usepackage{multicol}
\usepackage{multirow}
\usepackage{amsbsy}
\usepackage{array, caption, floatrow, tabularx, makecell, booktabs}%
\usepackage{animate}
\usepackage{floatrow}
\usepackage{etoolbox}
\usepackage{float}

\iccvfinalcopy 

\def\iccvPaperID{4121} 

\ificcvfinal\fi

\usepackage{xcolor}
\usepackage[export]{adjustbox}

\usepackage{url}
\usepackage{amsthm}
\usepackage{color}
\usepackage{subcaption}
\usepackage{booktabs}
\usepackage{arydshln}

\def\figref#1{Fig.~\ref{#1}}

\def\tabref#1{Table~\ref{#1}}

\DeclareGraphicsExtensions{.pdf,.jpg}

\graphicspath{ {images/}{syntheticExp/} {final_images/channel_gated/} {final_images/} {paper_images/} }

\usepackage[pagebackref=true,breaklinks=true,letterpaper=true,colorlinks,bookmarks=false]{hyperref}

\def\iccvPaperID{4121} 

\ificcvfinal\fi
\begin{document}

\title{Interactive Sketch \& Fill: Multiclass Sketch-to-Image Translation}

\author{
Arnab Ghosh$^{1}$ \hspace{8mm} Richard Zhang$^{2}$ \hspace{8mm} Puneet K. Dokania$^{1}$ \\
Oliver Wang$^{2}$ \hspace{8mm} Alexei A. Efros$^{2,3}$ \hspace{8mm} Philip H. S. Torr$^{1}$ \hspace{8mm} Eli Shechtman$^{2}$ \\
\\
$^{1}$University of Oxford \hspace{15mm} $^{2}$Adobe Research \hspace{15mm} $^{3}$UC Berkeley \\
}



\twocolumn[{%
\renewcommand\twocolumn[1][]{#1}%
\maketitle
\begin{center}
    \centering
    \includegraphics[width=1.\linewidth]{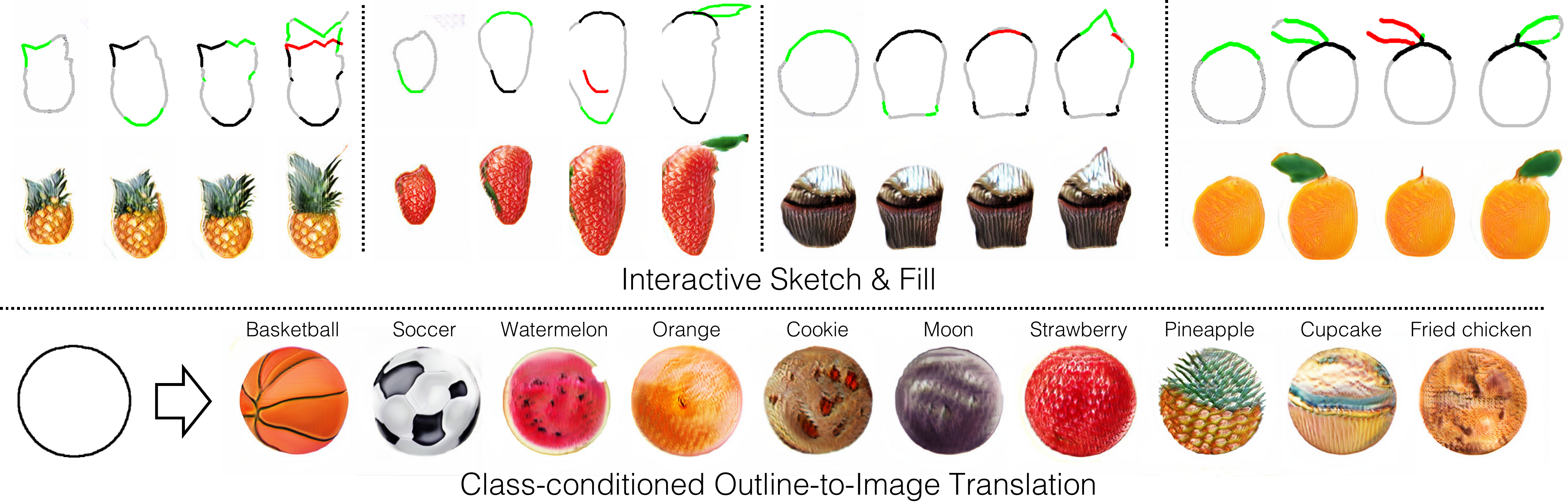}
    
    \captionof{figure}{({\bf Top}) Given a user created incomplete object outline (first row), our model estimates the complete shape and provides this as a recommendation to the user (shown in gray), along with the final synthesized object (second row). These estimates are updated as the user adds (\textcolor{green}{green}) or removes (\textcolor{red}{red}) strokes over time -- previous edits are shown in black.
    ({\bf Bottom}) This generation is class-conditioned, and our method is able to generate distinct multiple objects for the same outline (\eg `circle') by conditioning the generator on the object category.\label{fig:teaser}
    }
    \vspace{1em}
\end{center}%
}]
\begin{abstract}
We propose an interactive GAN-based sketch-to-image translation method that helps novice users create images of simple objects.
As the user starts to draw a sketch of a desired object type, the network interactively recommends plausible completions, and shows a corresponding synthesized image to the user. This enables a feedback loop, where the user can edit their sketch based on the network's recommendations, visualizing both the completed shape and final rendered image while they draw.
In order to use a single trained model across a wide array of object classes, we introduce a gating-based approach for class conditioning, which allows us to generate distinct classes without feature mixing, from a single generator network. 
\end{abstract}

\section{Introduction}
%
Conditional GAN-based image translation \cite{isola2016image2image,sangkloy2017scribbler,zhu2017unpaired} models have shown remarkable success at taking an abstract input, such as an edge map or a semantic segmentation map, and translating it to a real image. Combining this with a user interface allows a user to quickly create images in the target domain. 
However, such interfaces for object creation require the entire edge or label map as input, which is a challenging task as users typically create drawings \emph{incrementally}.
Furthermore, completing a line drawing without any feedback may prove difficult for many, as untrained practitioners generally struggle at free-hand drawing of accurate proportions of objects and their parts~\cite{cohen1997can}, 3D shapes and perspective~\cite{schmidt2009expert}. 
As a result, it is much easier with current interactive image translation methods to obtain realistic looking images by editing \emph{existing} images~\cite{dekel2018sparse,portenier2018faceshop} rather than creating images from scratch. 

We propose a new GAN-based interactive image generation system for drawing objects from scratch that: 1) generates full images given {\em partial} user strokes (or sketches); 2) serves as a \emph{recommender system} that suggests or helps the user \emph{during} their creative process to help them generate a desired image; and 3) uses a single conditional GAN model for {\em multiple} image classes, via a gating-based conditioning mechanism. Such a system allows for creative input to come from the user, while the challenging task of getting exact object proportions correct is left to the model, which constantly predicts a plausible completion of the user's sketch (\figref{fig:teaser}). 

Unlike other related work, we use sparse object outlines/sketches/simplified-edges instead of dense edge maps as the user input
as these are closer to the lines that novice users tend to draw~\cite{cole2008people}. 
Our model first completes the user input and then generates an image conditioned on the completed shape. There are several advantages to this two-stage approach.
For one, we are able to give the artist feedback on the general object shape in our interactive interface (similar to ShadowDraw~\cite{lee2011shadowdraw}), allowing them to quickly refine higher level shape until it is satisfactory.
Second, we found that splitting completion and image generation to work better than going directly from partial outlines to images, as the additional intermediate supervision on full outlines/sketches breaks the problem into two easier sub-problems -- first recover the geometric properties of the object (shape, proportions) and then fill in the appearance (colors, textures).


For the second stage, we use a multi-class generator that is conditioned on a user supplied class label. This generator applies a gating mechanism that allows the network to focus on the important parts (activations) of the network specific to a given class. Such an approach allows for a clean separation of classes, enabling us to train a single generator and discriminator across \emph{multiple} object classes, thereby enabling a finite-size deployable model that can be used in multiple different scenarios. 

To demonstrate the potential of our method as an interactive tool for stroke-based image generation, we collect a new image dataset of ten simple object classes (pineapple, soccer, basketball, etc.) with white backgrounds. In order to stress test our conditional generation mechanism, six of the object classes have similar round shapes, which requires the network to derive texture information from the class conditioning. \figref{fig:gui} shows a short video of an interactive editing session using our system. Along with these simple objects, we also demonstrate the potential of our method on more complicated shapes, such as faces and shoes. Code and other details are available at our \href{https://arnabgho.github.io/iSketchNFill/}{website}.

\begin{figure*}[t]
    \centering  
    \begin{tabular}{cc}
    \animategraphics[autoplay,loop,width=.45\linewidth]{25}{images/gif/}{00001}{00266} &
    \animategraphics[autoplay,loop,width=.50\linewidth]{25}{images/gif_shadow/}{00001}{00096}
    \\
    \end{tabular}
        \caption{{\bf Video of our interface} We can see two versions of our interface. The left side shows how a user can quickly generate multiple objects using a few strokes, while the right side shows the utility of multimodal completions where the user can quickly explore different possible shape generations while drawing. Full video available at our \href{https://arnabgho.github.io/iSketchNFill/}{website}. { \textbf{Please view with Acrobat Reader.}}}\label{fig:gui}
\end{figure*}

\section{Related Work}

\paragraph{Interactive Generation} Interactive interfaces for freehand drawing go all the way back to Ivan Sutherland's Sketchpad~\cite{sutherland64}.  The pre-deep work most related to us, ShadowDraw~\cite{lee2011shadowdraw}, introduced the concept of generating multiple shadows for novice users to be able to draw sketches. PhotoSketcher \cite{eitz2011photosketcher} introduces a retrieval based method for obtaining real images from sketches. 
More recently, deep recurrent networks have been used to generate sketches~\cite{ha2017neural,ganin2018synthesizing}. Sketch-RNN~\cite{ha2017neural} provides a completion of partial strokes, with the advantage of intermediate stroke information via the Quickdraw dataset at training time. SPIRAL \cite{ganin2018synthesizing} learns to generate digits and faces using a reinforcement learning approach.
Zhu et al.~\cite{zhu2016generative} train a generative model, and an optimization-based interface to generate possible images, given color or edge constraints. The technique is limited to a single class and does not propose a recommendation for the completion of the shape. SketchyGAN~\cite{chen2018sketchygan} also aimed at generating multi-class images but lacks interactive capability. In contrast to the above, our method provides interactive prediction of the shape and appearance to the user and supports multiple object classes.
\paragraph{Generative Modeling} Parametric modeling of an image distribution is a challenging problem. Classic approaches include autoencoders~\cite{hinton2006reducing,vincent2008extracting} and Boltzmann machines~\cite{smolensky1986information}. More modern approaches include autoregressive models~\cite{efros1999texture,van2016conditional}, variational autoencoders (VAEs)~\cite{kingma2013auto}, and generative adversarial networks (GANs). GANs and VAEs both learn mappings from a low-dimensional ``latent" code, sampled stochastically, to a high-dimensional image through a feedforward pass of a network. GANs have been successful recently~\cite{denton2015deep,radford2015unsupervised,arjovsky2017wgan}, and hybrid models feature both a learned mapping from image to latent space as well as adversarial training~\cite{donahue2016adversarial,dumoulin2016adversarially,larsen2016vaegan,chen2016infogan}.

\paragraph{Conditioned Image Generation} The methods described above can be conditioned, either by a low-dimensional vector (such as an object class, or noise vector), a high-dimensional image, or both. Isola et al.~\cite{isola2016image2image} propose ``pix2pix", establishing the general usefulness of conditional GANs for image-to-image translation tasks. However, they discover that obtaining multimodality by injecting a random noise vector is difficult, a result corroborated in~\cite{mathieu2015deep,pathak2016context,zhu2017toward}.
This is an example of mode collapse~\cite{goodfellow2016nips}, a phenomenon especially prevalent in image-to-image GANs, as the generator tends or ignore the low-dimensional latent code in favor of the high-dimensional image.
Proposed solutions include layers which better condition the optimization, such as Spectral Normalization~\cite{zhang2018self,miyato2018spectral}, modifications to the loss function, such as WGAN~\cite{arjovsky2017wasserstein,gulrajani2017improved} or optimization procedure~\cite{heusel2017gans}, or modeling proposals, such as MAD-GAN~\cite{ghosh2017multi} and MUNIT~\cite{huang2018multimodal}. 
One modeling approach is to add a predictor from the output to the conditioner, to discourage the model from ignoring the conditioner. This has been explored in the classification setting in Auxiliary-Classifier GAN (ACGAN)~\cite{odena2016conditional} and regression setting with InfoGAN~\cite{chen2016infogan} and ALI/BiGAN (``latent regressor" model)~\cite{dumoulin2016adversarially,donahue2016adversarial}, and is one half of BicycleGAN model~\cite{zhu2017toward}. We explore a complementary approach of architectural modification via gating.




\noindent \textbf{Gating Mechanisms}
Residual networks~\cite{he2016deep}, first introduced for image classification~\cite{krizhevsky2012imagenet}, have made extremely deep networks viable to train. Veit et al.~\cite{veit2016residual} find that the skip connection in the architecture enables test-time removal of blocks. Follow-up work~\cite{veit2018adaptive} builds in block removal during training time, with the goal of subsets of blocks specializing to different categories. Inspired by these results, we propose the use of gating for image generation and provide a systematic analysis of gating mechanisms.

The adaptive instance normalization (AdaIn) layer has similarly been used in arbitrary style transfer~\cite{huang2017arbitrary} and image-to-image translation~\cite{huang2018multimodal}, and Feature-wise Linear Modulation (FiLM)~\cite{perez2017film}. Both methods scale and shift feature distributions, based on a high-dimensional conditioner, such as an image or natural language question. Gating also plays an important role in sequential models for natural language processing: LSTMs \cite{hochreiter1997long} and GRU \cite{cho2014learning}. Similarly, concurrent work \cite{karras2018style}, \cite{park2019semantic} use a AdaIN-style network to modulate the generator parameters.

\begin{figure}[t]
    \centering  
    \includegraphics[width=\linewidth]{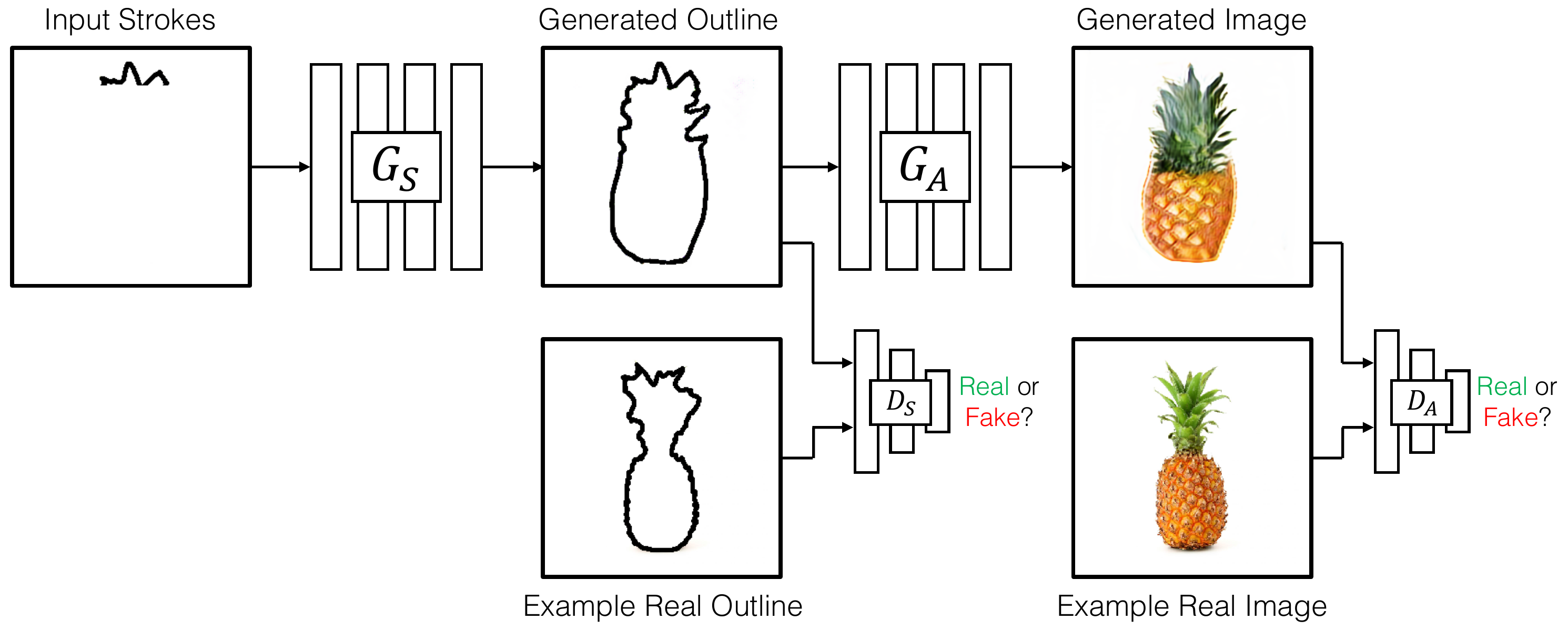} 
    \\
    \vspace{-4mm}
        \caption{{\bf Our two-stage approach} First, we complete a partial sketch using the shape generator $G_S$. Then we translate the completed sketch into an image using the appearance generator $G_A$. Both generators are trained with their respective discriminators $D_S$, and $D_A$.
        {}}\label{fig:SketchFillNet}
\end{figure}

\begin{figure*}[t]
    \centering
    \includegraphics[width=\linewidth]{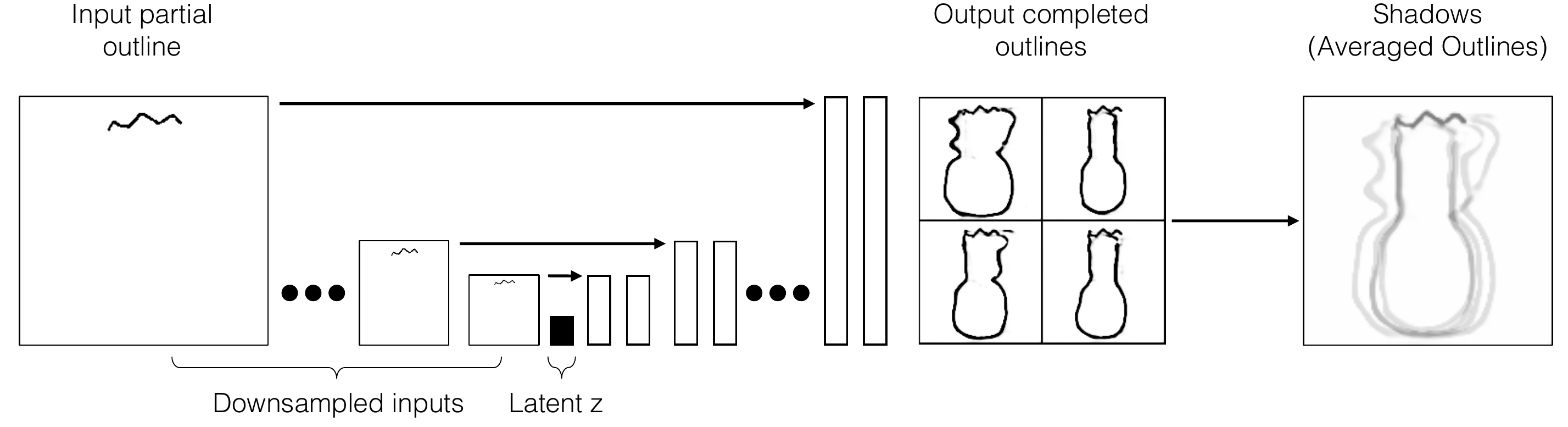}
    \vspace{-8mm}
    \caption{\textbf{First stage (Shape Generator)} To achieve multi-modal completions, the shape generator is designed using inspiration from non-image conditional model \cite{mescheder2018training} with the conditioning input provided at multiple scales, so that the generator network doesn't ignore the partial stroke conditioning.
    }\label{fig:SketchNet}
\end{figure*}

\label{sec:methods}

\vspace{-4mm}
\section{Method}
We decouple the problem of interactive image generation into two stages: object shape completion from sparse user sketches, and appearance synthesis from the completed shape. More specifically, as illustrated in \figref{fig:SketchFillNet} we use the Shape Generator $G_S$ for the automatic shape (outline/sparse-sketch/simplified-edge) generation and the Appearance Generator $G_A$ for generating the final image as well as the adversary discriminators $D_S$ and $D_A$. Example usage is shown in our user interface in \figref{fig:gui}.

\subsection{Shape completion}
\label{sec:shape}
\begin{figure}[t]
\centering
\begin{tabular}{*{4}{c@{\hspace{3px}}}}
    \frame{\includegraphics[width=.22\linewidth]{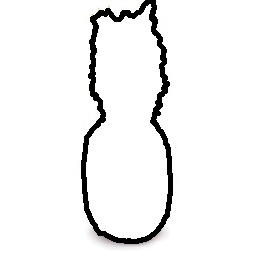}} &
    \frame{\includegraphics[width=.22\linewidth]{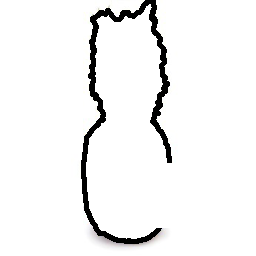}} &
    \frame{\includegraphics[width=.22\linewidth]{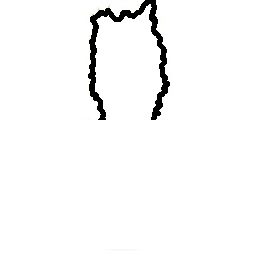}} &
    \frame{\includegraphics[width=.22\linewidth]{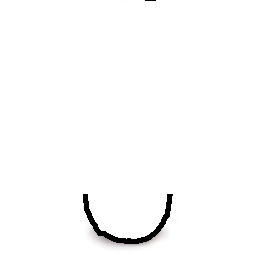}}\\
    Outline &
    \multicolumn{3}{c}{Simulated Partial Inputs}
    \\
\end{tabular} \\
\vspace{-3mm}
    \caption{\textbf{Simulated Inputs} Three sizes of occluders were used to simulate partial outlines.}
    \label{fig:autocomplete_data_generation}
\end{figure}

\begin{figure}[t]
\centering
\begin{tabular}{*{4}{c@{\hspace{3px}}}}
    \frame{\includegraphics[width=.22\linewidth]{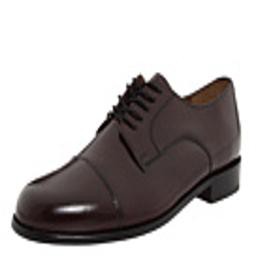}} &
    \frame{\includegraphics[width=.22\linewidth]{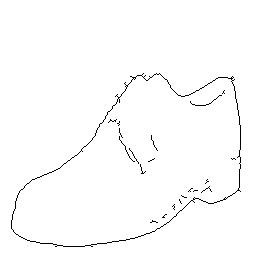}} &
    \frame{\includegraphics[width=.22\linewidth]{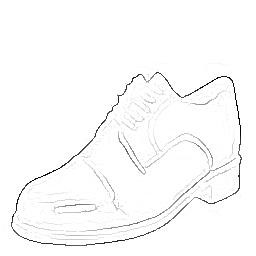}} &
    \frame{\includegraphics[width=.22\linewidth]{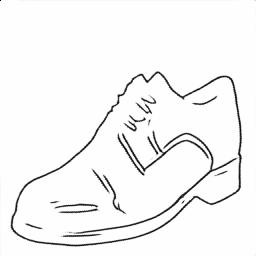}}
    \\
    
\end{tabular} \\
    \caption{\textbf{Simplified Edges} The 2nd edgemap is obtained using the technique of \cite{isola2016image2image}, while the 3rd is the intermediate edgemap using \cite{li2019im2pencil} and further simplified using \cite{simo2016learning} which looks closer to what a human would sketch. }
    \label{fig:simplified_edges}
    \vspace{-3mm}
\end{figure}

The shape completion network $G_S$ should provide the user with a visualization of its completed shape(s), based on the user input, and should keep on updating the suggested shape(s) interactively. 
We take a data-driven approach for this whereby, to train the network, we simulate partial strokes (or inputs) by removing random square patches from the full outline/ full sparse sketch/ full simplified edges. 
The patches are of three sizes (64$\times$64, 128$\times$128, 192$\times$192) and placed at a random location in the image of size 256$\times$256 (see \figref{fig:autocomplete_data_generation} for an example). To extend the technique beyond outlines and generate more human-like sketches, we adopt the multistage procedure depicted in \figref{fig:simplified_edges}. We refer to these generated sketches as ``simplified edges''.
We automatically generate data in this manner, creating a dataset where for a given full outline/sketch or a simplified edge-map, 75 different inputs are created.
The model, shown in \figref{fig:SketchFillNet}, is based on the architecture used for non-image conditional generations in \cite{mescheder2018training}. We modify the architecture such that the conditioning input is provided to the generator and discriminator at multiple scales as shown in \figref{fig:SketchNet}. This makes the conditioning input an active part of the generation process and helps in producing multimodal completions.

\vspace{-4mm}
\begin{figure*}[t]
    \centering
    \includegraphics[width=1.\linewidth]{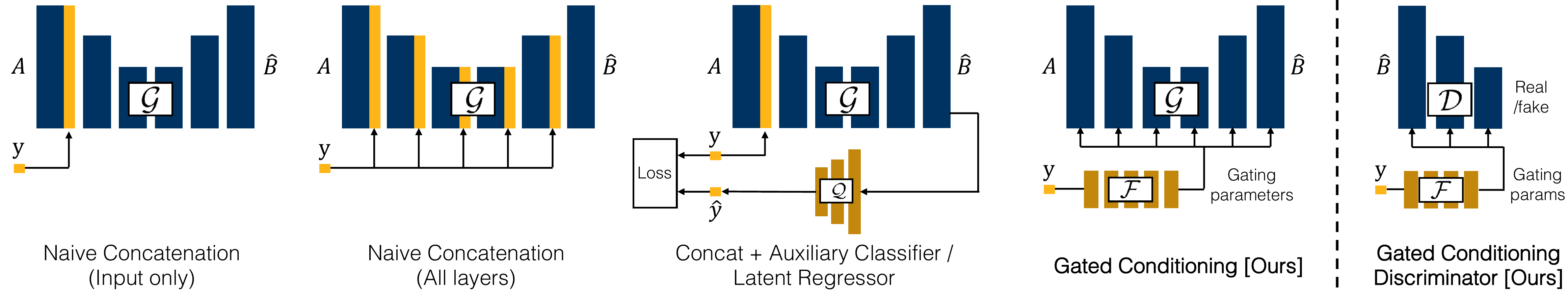}
    \vspace{-6mm}
    \caption{
    {\bf Conditioning variants for the Appearance Generator} Our model uses gating on all the residual blocks of the generator and the discriminator, other forms of conditioning such as (naive concatenation in input only, all layers, AC-GAN like latent regressor \cite{odena2016conditional}) are evaluated as well. \label{fig:arch-gate1}
    }
\end{figure*}

\begin{figure*}[t]
    \centering
    \includegraphics[width=1.\linewidth]{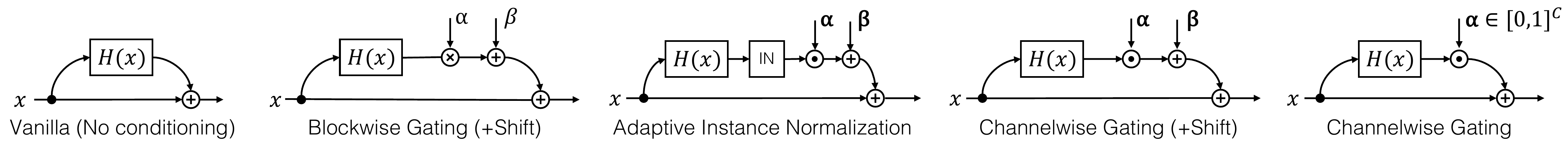}
    \vspace{-4mm}
    \caption{
    {\bf Injecting conditioning with modified residual layers} {\bf (Left)} A ``vanilla" residual block without conditioning applies a residual modification to the input tensor. {\bf (Mid-left)} The $\mathcal{H}(X)$ block is softly-gated by scalar parameter $\alpha$ and shift $\beta$. {\bf (Mid)} Adaptive Instance Normalization~\cite{huang2017arbitrary} applies a channel-wise scaling and shifting after an instance normalization layer. {\bf (Mid-right)} Channel-wise gating adds restrictions to the range of $\mbox{\boldmath $\alpha$}$. {\bf (Right)} We find that channel-wise gating (without added bias) produces the best results empirically.\label{fig:arch-gate2}
    \vspace{-2mm}
    }
\end{figure*}

\vspace{4mm}
\subsection{Appearance synthesis}
\label{sec:appearance}
An ideal interactive sketch-to-image system  should be able to generate multiple different image classes with a single generator. 
Beside memory and time considerations (avoiding loading/using a separate model per class, reducing overall memory), a single network can share features related to outline recognition and texture generation that are common across classes, which helps training with limited examples per class. 

As we later show, class-conditioning by concatenation can fail to properly condition the network about the class information in current image translation networks~\cite{isola2016image2image,zhu2017toward}.
To address this, we propose an effective soft gating mechanism, shown in~\figref{fig:arch-gate1}.
Conceptually, our network consists of a small external gating network that is conditioned on the object class (encoded as a 1-hot vector).
The gating network outputs parameters that are used to modify the features of the main generator network.
%
Given an input feature tensor $X_l$, ``vanilla'' ResNet~\cite{he2016deep} maps it to
\begin{equation}
X_{l+1} = X_l+\mathcal{H}_l(X_l).
\end{equation}
Changes in resolution are obtained by upsampling before or downsampling after the residual block.
Note that we omit $l$ subscript from this point forward to reduce clutter.
Our gating network augments this with a predicted scalar $\alpha$ for each layer of the network using a learned network $\mathcal{F}({\bf y})$, where ${\bf y}$ is the conditioning vector:
\begin{equation}
X + \alpha \; \mathcal{H}(X), \text{where } \alpha \in [0,1]
\end{equation}

If the conditioning vector ${\bf y}$ has no use for a particular block, it can predict $\alpha$ close to zero and effectively switch off the layer.
During training, blocks within the main network can transform the image in various ways, and $\mathcal{F}$ can modulate such that the most useful blocks are selected. 
Unlike previous feature map conditioning methods such as AdaIn~\cite{ulyanovinstance}, we apply gating to \emph{both} the generator and discriminator. 
This enables the discriminator to select blocks which effectively judge whether generations are real or fake, conditioned on the class input.
Some blocks can be shared across regions in the conditioning vector, whereas other blocks can specialize for a given class.

A more powerful method is to apply this weighting channel-wise using a vector {\boldmath$\alpha$}: 
\begin{align}
X + \mbox{\boldmath $\alpha$} \odot \mathcal{H}(X), \text{where } \mbox{\boldmath $\alpha$} \in [0,1]^c, 
\end{align}
where $\odot$ represents channel-wise multiplication. This allows specific channels to be switched ``on" or ``off", providing additional degrees of freedom.
We found that this channelwise approach for gating provides the strongest results. 
AdaIn describes the case where an Instance Normalization~\cite{ulyanovinstance} (IN) operation is applied before scaling and shifting the feature distribution.
We constrain each element of {\boldmath $\alpha$} and {\boldmath $\beta$} in $[-1, 1]$.
We additionally explored incorporating a bias term after the soft-gating, either block-wise using a scalar $\beta \in [-1,1]$ per layer, or channel-wise using a vector $\mbox{\boldmath $\beta$} \in [-1, 1]^c$ per layer but we found that they did not help much, and so we leave them out of our final model. Refer \figref{fig:arch-gate2} for pictorial representation of various gatings.

Finally, we describe our network architecture, which utilizes the gated residual blocks described above.
We base our architecture on the proposed residual \textbf{Encoder-Decoder} model from MUNIT~\cite{huang2018multimodal}.
This architecture is comprised of 3 \texttt{conv} layers, 8 residual blocks, and 3 \texttt{up-conv} layers. The residual blocks have 256 channels. 
First, we deepen the network, based on the principle that deeper networks have more valid disjoint, partially shared paths~\cite{veit2016residual}, and add 24 residual blocks. 
To enable the larger number of residual blocks, we drastically reduce the width to 32 channels for every layer. 
We refer to this network as \textbf{SkinnyResNet}. 
Additionally, we found that modifying the downsampling and upsampling blocks to be residual connections as well improved results, and also enables us to apply gating to {\em all} blocks. 
When gating is used, the gate prediction network, $\mathcal{F} ({\bf y})$,  
is also designed using residual blocks. Additional architecture details are in the supplementary material.

\begin{figure*}[t]
\begin{tabular}{cc}
      \resizebox{0.5\linewidth}{!}{
\begin{tabular}{*{5}{c@{\hspace{3px}}}}
    \frame{\includegraphics[width=.12\linewidth]{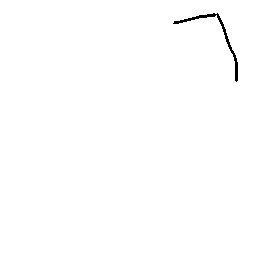}} &
    \frame{\includegraphics[width=.12\linewidth]{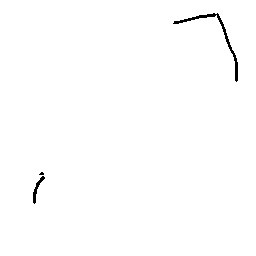}} & 
    \frame{\includegraphics[width=.12\linewidth]{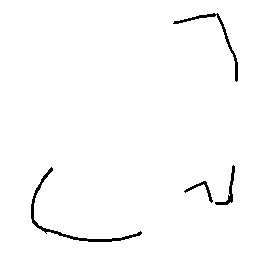}}&
    \frame{\includegraphics[width=.12\linewidth]{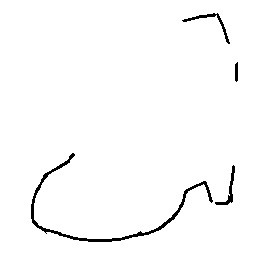}} &
    \\
    \frame{\includegraphics[width=.12\linewidth]{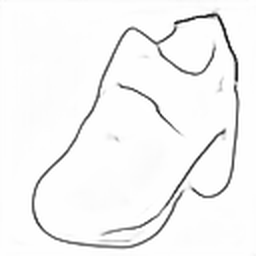}} &
    \frame{\includegraphics[width=.12\linewidth]{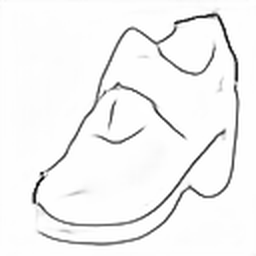}} & 
    \frame{\includegraphics[width=.12\linewidth]{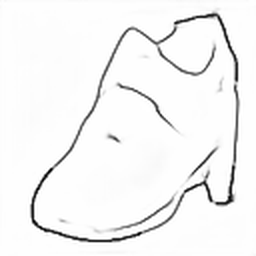}}&
    \frame{\includegraphics[width=.12\linewidth]{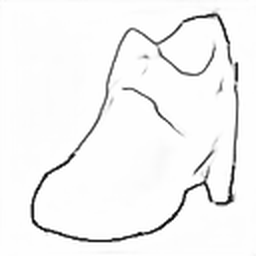}} &
   
    \\
    \frame{\includegraphics[width=.12\linewidth]{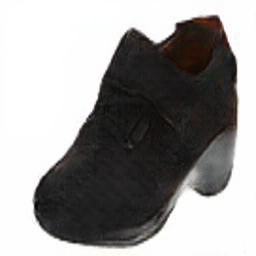}} &
    \frame{\includegraphics[width=.12\linewidth]{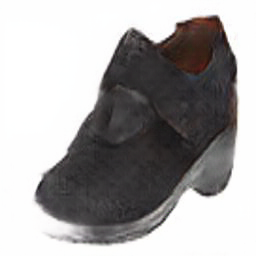}} & 
    \frame{\includegraphics[width=.12\linewidth]{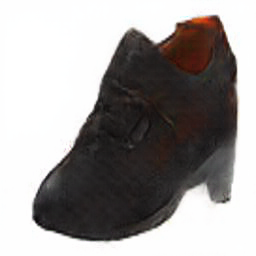}}&
    \frame{\includegraphics[width=.12\linewidth]{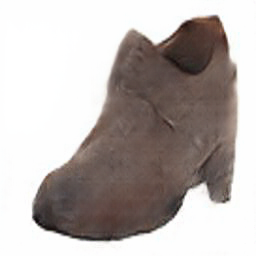}} &
    
    \\
\end{tabular}
    }
     & 
    \resizebox{0.5\linewidth}{!}{
\begin{tabular}{*{5}{c@{\hspace{3px}}}}
    \frame{\includegraphics[width=.12\linewidth]{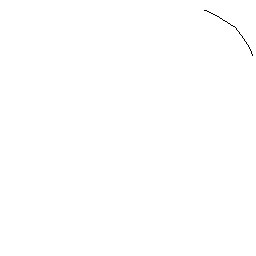}} &
    \frame{\includegraphics[width=.12\linewidth]{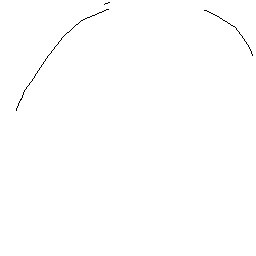}} & 
    \frame{\includegraphics[width=.12\linewidth]{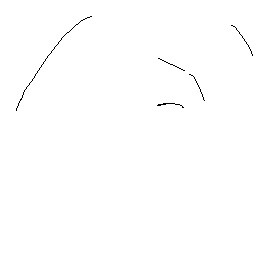}}&
    \frame{\includegraphics[width=.12\linewidth]{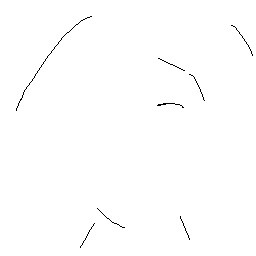}} &
    \\
    \frame{\includegraphics[width=.12\linewidth]{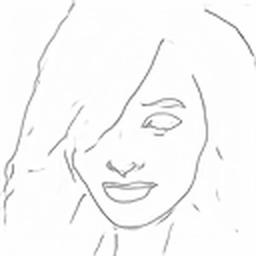}} &
    \frame{\includegraphics[width=.12\linewidth]{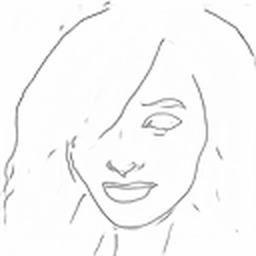}} & 
    \frame{\includegraphics[width=.12\linewidth]{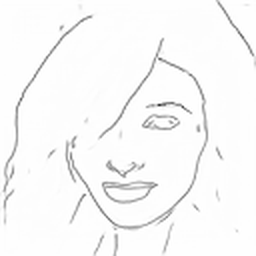}}&
    \frame{\includegraphics[width=.12\linewidth]{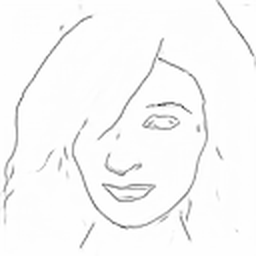}} &
   
    \\
    \frame{\includegraphics[width=.12\linewidth]{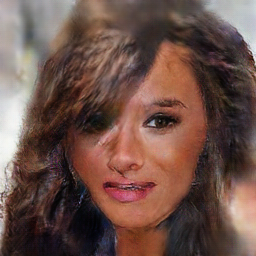}} &
    \frame{\includegraphics[width=.12\linewidth]{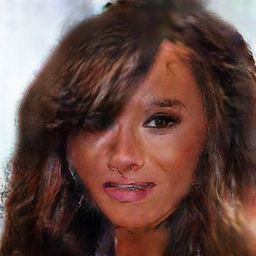}} & 
    \frame{\includegraphics[width=.12\linewidth]{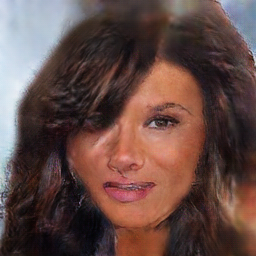}}&
    \frame{\includegraphics[width=.12\linewidth]{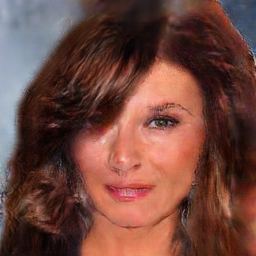}} &
    
    \\
\end{tabular}
} 
     
\end{tabular}
\vspace{-2mm}
    \caption{\textbf{Example Sketch \& Fill Progression.} The \textbf{first row} represents the progressive addition of new strokes on the canvas, the \textbf{second row} shows the autocompleted sketch, and the \textbf{third row} is the final generated image. 
    As the sparse strokes are changed by the user, the completed shape and generated image evolve as well. Note that changing a stroke locally produces coherent changes in other parts of the image.
    \vspace{-4mm}
    }
    \label{fig:autocomplete_generate_sketches}
\end{figure*}

\begin{figure*}[t]
    \centering
    \includegraphics[width=.9\linewidth]{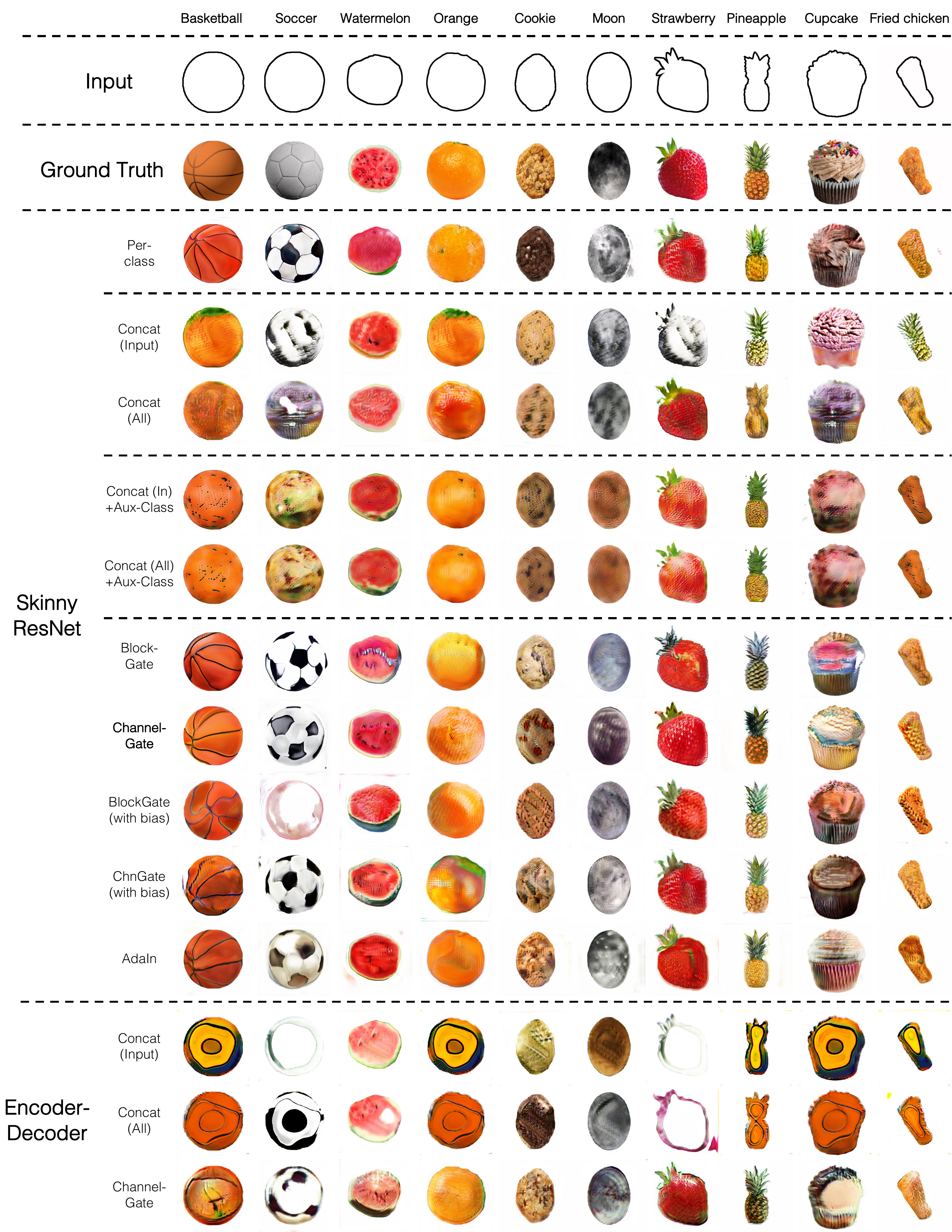}
    \caption{{\bf Conditioning injection comparison.} We show results across methods on the outline$\rightarrow$image task using the \textbf{SkinnyResNet} architecture. Naive Concatenation \textbf{Concat} often confuses classes, such as oranges and basketballs, while gating mechanisms such as the \textbf{ChannelGate} method succeed. The gating method also improves results for the \textbf{EncoderDecoder} architecture. \label{fig:alg_comp} }
    \vspace{-4mm}
\end{figure*}

\vspace{-4mm}
\section{Experiments}
\label{sec:experiments}
We first compare our 2 step approach for interactive image generation on existing datasets such as the UTZappos Shoes dataset \cite{yu2014fine} and CelebA-HQ \cite{karras2017progressive}. State-of-the-art techniques such as pix2pixHD \cite{Wang_2018_CVPR} are used to generate the final image from the autocompleted sketches. We finally evaluate our approach on a multi-class dataset that we collected to test our proposed gating mechanism.


\begin{table}[t]
\resizebox{1.\linewidth}{!} {
    \centering
        \begin{tabular}{l c}
        \toprule
        \textbf{Trained task} & \textbf{FID} \\ \midrule
        \textbf{Faces}\\ \hline
        Partial Simplified Edges $\rightarrow$ Image & 383.02 \\
        Partial Simplified Edges $\rightarrow$ Simplified Edges $\rightarrow$ Image & 374.67 \\
        \hline
        \textbf{Shoes}\\ \hline
        Partial Simplified Edges $\rightarrow$ Image & 170.45 \\
        Partial Simplified Edges $\rightarrow$ Simplified Edges $\rightarrow$ Image & 154.32 \\
        \bottomrule 
        \end{tabular}
        }
    \vspace{-2mm}
    \caption{\label{table:2step_eval_single_class} \textbf{Single-class generation, 2-stage vs 1-stage}. We evaluate the result quality from different task pipelines.
    \vspace{-4mm}
    }
\end{table}
\vspace{-2mm}
\subsection{Single Class Generation}
\paragraph{Datasets} We use the edges2shoes\cite{isola2016image2image}, CelebA-HQ\cite{karras2017progressive} datasets to test our method on single class generation. 
We simplify the edges to attempt to more closely resemble how humans would draw strokes by first using the preprocessing code of \cite{li2019im2pencil} further reducing the strokes with a sketch simplification network \cite{simo2016learning}.
\paragraph{Architecture} We use the architecture described in Section \ref{sec:shape} for shape completion. In this case, each dataset only contains a single class, so we can use an off-the-shelf network, such as pix2pixHD~\cite{wang2017high} for rendering.

\paragraph{Results} As seen in \figref{fig:autocomplete_generate_sketches}, our 2 step technique allows us to complete the simplified edge maps from the partial strokes and also generate realistic images from the autocompleted simplified edges.
Table \ref{table:2step_eval_single_class} also demonstrates, across two datasets (faces and shoes), that using a 2 step procedure produces stronger results than mapping directly from the partial sketch to the completed image.

\begin{table}[t]
    \centering
        \begin{tabular}{l c}
        \toprule
        \textbf{Trained task} & \textbf{Avg Acc} \\ \midrule
        Partial edges $\rightarrow$ Image & 73.12 \% \\
        Partial outline $\rightarrow$ Image & 88.74 \% \\
        Partial outline $\rightarrow$ Full outline $\rightarrow$ Image [Ours] & \textbf{97.38\%}  \\
        \bottomrule 
        \end{tabular}
    \vspace{-2mm}
    \caption{\label{table:2step_eval} \textbf{Multi-class generation, 2-stage vs 1-stage}. We evaluate the result quality from different task pipelines. Accuracy is computed by a fixed, pretrained classification network, on the resulting images.
    \vspace{-1mm}
    }
\end{table}

\begin{figure}[t]
\centering
\begin{tabular}{*{6}{c@{\hspace{3px}}}}
    \frame{\includegraphics[width=.15\linewidth]{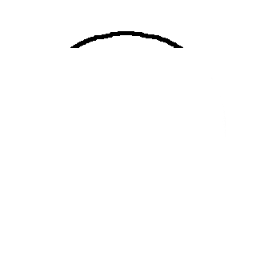}} &
    \frame{\includegraphics[width=.15\linewidth]{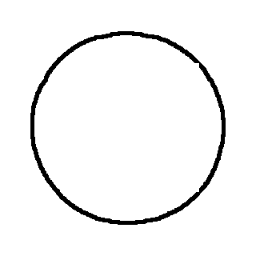}} &
    \frame{\includegraphics[width=.15\linewidth]{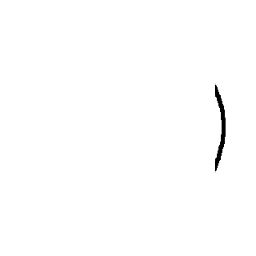}} & 
    \frame{\includegraphics[width=.15\linewidth]{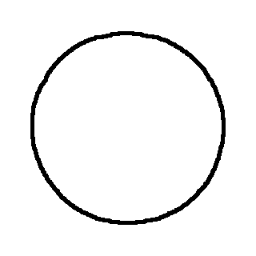}}&
    \frame{\includegraphics[width=.15\linewidth]{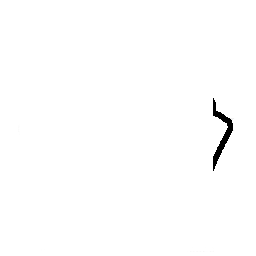}} &
    \frame{\includegraphics[width=.15\linewidth]{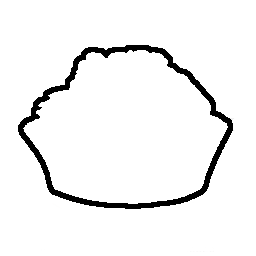}}
    \\
    
    \frame{\includegraphics[width=.15\linewidth]{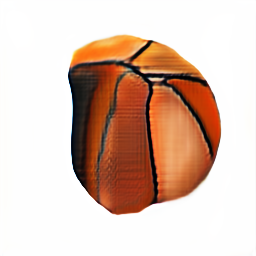}} &
    \frame{\includegraphics[width=.15\linewidth]{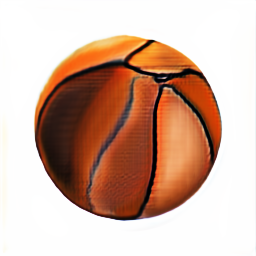}} &
    \frame{\includegraphics[width=.15\linewidth]{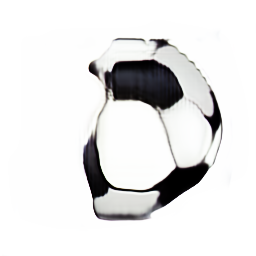}} & 
    \frame{\includegraphics[width=.15\linewidth]{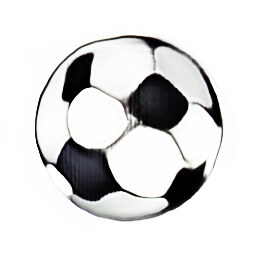}}&
    \frame{\includegraphics[width=.15\linewidth]{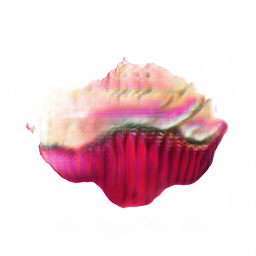}} &
    \frame{\includegraphics[width=.15\linewidth]{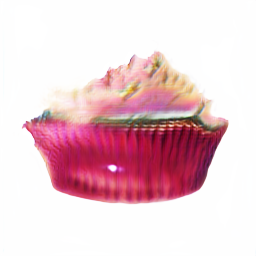}}
    \\
    
\end{tabular}
    \caption{\textbf{Directly mapping from partial outline to image} Our proposed system uses a 2-stage approach, using a completed edge map as an intermediate. Here, we show results when directly mapping from the partial outline to the image. When the outline is well-defined, the network can generate realistic images. However, when the outline is sparse, the network struggles with the geometry.}
    \label{fig:ablation_partial_full_outline}
    \vspace{-3mm}
\end{figure}

\begin{figure}[t]
\centering
\resizebox{1.0\linewidth}{!}{
\begin{tabular}{*{5}{c@{\hspace{3px}}}}
    \frame{\includegraphics[width=.2\linewidth]{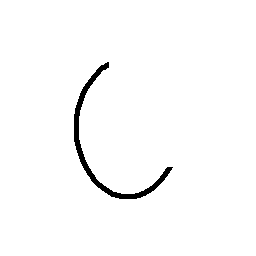}} &
    \frame{\includegraphics[width=.2\linewidth]{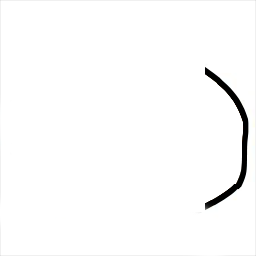}} & 
    \frame{\includegraphics[width=.2\linewidth]{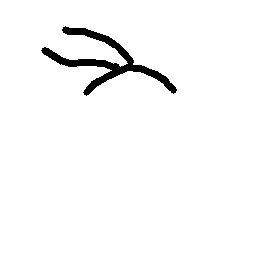}}&
    \frame{\includegraphics[width=.2\linewidth]{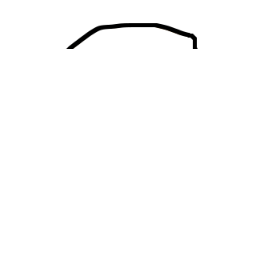}} &
    \frame{\includegraphics[width=.2\linewidth]{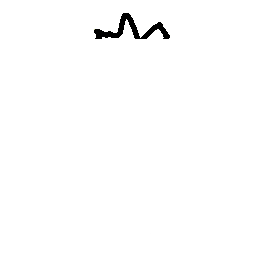}}
    \\
    
    \frame{\includegraphics[width=.2\linewidth]{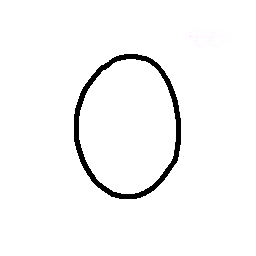}} &
    \frame{\includegraphics[width=.2\linewidth]{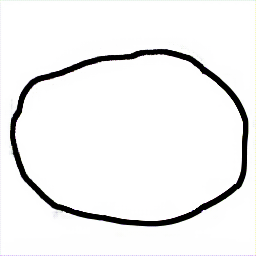}} & 
    \frame{\includegraphics[width=.2\linewidth]{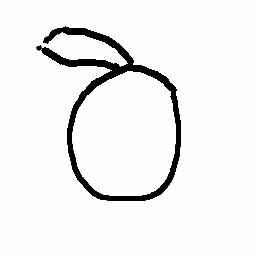}}&
    \frame{\includegraphics[width=.2\linewidth]{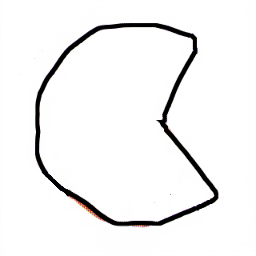}} &
    \frame{\includegraphics[width=.2\linewidth]{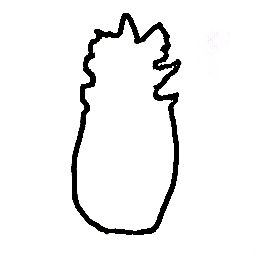}}
    \\

    \frame{\includegraphics[width=.2\linewidth]{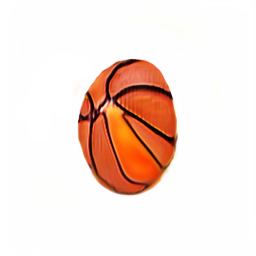}} &
    \frame{\includegraphics[width=.2\linewidth]{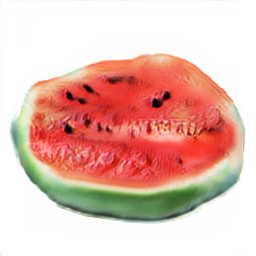}} & 
    \frame{\includegraphics[width=.2\linewidth]{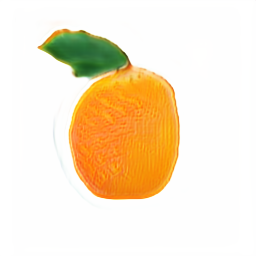}}&
    \frame{\includegraphics[width=.2\linewidth]{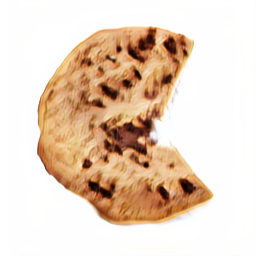}} &
    \frame{\includegraphics[width=.2\linewidth]{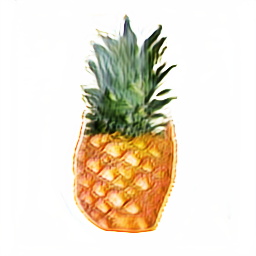}}
    \\
\end{tabular}
}
    \caption{\textbf{Multiclass Sketch \& Fill results}
    A few input strokes (first row) are enough to automatically complete the class specific outlines (second) and appearance (last).
    \vspace{-3mm}
    }
    \label{fig:autocomplete_generate}
\end{figure}


\begin{table}[]
  \centering
   \resizebox{1.\linewidth}{!} {
  \setlength{\tabcolsep}{6pt}
  \begin{tabular}{l c c c c}
  \toprule
    \multirow{3}{*}{\textbf{Method}} & \multicolumn{2}{c}{ {\bf SkinnyResNet}} & \multicolumn{2}{c}{ {\bf EncDec}} \\ \cmidrule(l){2-3} \cmidrule(l){4-5}
	& Class. & AMT Fool. & Class. & AMT Fool. \\
	& Acc [\%] & Rate [\%] & Acc [\%] & Rate [\%] \\ \midrule
    Ground truth & 100.0 & 50.0 & 100.0 & 50.0 \\ \midrule
    1 gen/class & \textbf{\textit{97.0}} & 17.7$\pm$1.46 & -- & -- \\ \midrule
    Concat (In)	& 62.6 & 15.0$\pm$1.4 & 39.2 & 7.5$\pm$1.06 \\ 
    Concat (All) & 64.5 & 15.3$\pm$1.41 & 51.4 & 5.4$\pm$0.88 \\ \midrule
    Cat(In)+Aux-Class & 65.6 & 14.5$\pm$1.5 & -- & -- \\ 
    Cat(All)+Aux-Class & 67.0 & 19.7$\pm$1.42 & -- & --\\ \midrule
    BlockGate(+bias) & 89.6 & 19.6$\pm$1.34 & -- & --\\ 
    BlockGate & {\bf 99.6} & 17.3$\pm$1.61 & -- & --\\ 
    AdaIn & 94.5 & 14.9$\pm$1.47 & -- & --\\ 
    ChanGate(+bias) & 94.1 & 14.8$\pm$1.43 & -- & --\\ 
    ChanGate & \textbf{\textit{97.0}} & {\bf 23.4$\pm$1.99} & 92.7 & 14.1$\pm$1.48 \\ 
	\hline
	\vspace{-5mm}
	  \caption{\small {\bf Accuracy vs Realism on Multiclass Outline$\rightarrow$Image task.} We measure generation accuracy with a pretrained network. We measure realism using the real vs. fake judges from AMT. Higher is better for both. Our SkinnyResNet architecture outperforms the Encoder-Decoder network, inspired by MUNIT~\cite{huang2018multimodal}. We perform a thorough ablation on our architecture and find that channel-wise gating achieves high accuracy and higher realism.
  \vspace{-5mm}
  }\label{fig:acc_vs_real}
	\end{tabular} 
	}
\end{table}

  

\subsection{Multi-Class Generation} 
\paragraph{Datasets} To explore the efficacy of our full pipeline, we introduce a new outline dataset consisting of 200 images (150 train, 50 test) for each of 10 classes -- basketball, chicken, cookie, cupcake, moon, orange, soccer, strawberry,  watermelon and pineapple. All the images have a white background and were collected using search keywords on popular search engines.
In each image, we obtain rough outlines for the image. We find the largest blob in the image after thresholding it into a black and white image. We fill the interior holes of the largest blob and obtain a smooth outline using the Savitzky–Golay filter~\cite{savitzky1964smoothing}.

\paragraph{Architecture} For the shape completion, we use the architecture in Section \ref{sec:shape}. For class-conditioned image generation, test the gated architectures in Section \ref{sec:appearance}.
\vspace{-4mm}
\paragraph{Results}
In order to test the fidelity of the automatically completed shapes, we evaluate the accuracy of a trained classifier on being able to correctly label a particular generation.
We first test in Table \ref{table:2step_eval} that our 2 stage technique is better than 1 step generation.
We evaluate the results on the multi-class outline to image generations on two axes: adherence to conditioning and realism. We first test the conditioning adherence -- whether the network generates an image of the correct class.
Off-the-shelf networks have been previously used to evaluate colorizations~\cite{zhang2016colorful}, street scenes~\cite{isola2016image2image, wang2017high}, and ImageNet generations~\cite{salimans2016improved}. 
We take a similar approach and fine-tune a pretrained InceptionV3 network~\cite{szegedy2016rethinking} for our 10 classes. 
The generations are then tested with this network for classification accuracy. Results are presented in Table~\ref{fig:acc_vs_real}.

To judge the generation quality, we also perform a ``Visual Turing test" using Amazon Mechanical Turk (AMT). Turkers are shown a real image, followed by a generated image, or vice versa, and asked to identify the fake. An algorithm which generates a realistic image will ``fool" Turkers into choosing the incorrect image. We use the implementation from~\cite{zhang2016colorful}.
Results are presented in Table~\ref{fig:acc_vs_real}, and qualitative examples are shown in Fig.~\ref{fig:alg_comp}.

\paragraph{Gating Architectures}
We compare our proposed model to the residual \textbf{Encoder-Decoder} model~\cite{huang2018multimodal}.
In addition, we compare our proposed gating strategy and {\bf SkinnyResNet} architecture to the following methods for  conditional image generation:

\begin{itemize}[noitemsep,leftmargin=12pt]
\item{\bf Per-class}: a single generator for each category; this is the only test setting with \textit{multiple} networks, all others train a single network
\item{\bf Concat (In)}: naive concatenation, input layer only
\item{\bf Concat (All)}: naive concatenation, all layers
\item{\bf Concat (In)+Aux-Class}: we add an auxiliary classifier, both for input-only and all layers settings
\item{\bf BlockGate(+Bias), BlockGate}: block-wise soft-gating, with and without a bias parameter
\item{\bf AdaIn}: Adaptive instance normalization
\item{\bf ChannelGate(+Bias), ChannelGate}: channel-wise soft-gating, with and without a bias parameter
\end{itemize}
\vspace{-2mm}



\vspace{2mm} \noindent \textbf{Does naive concatenation effectively inject conditioning?} In Fig.~\ref{fig:alg_comp}, we show a selected example from each of the 10 classes. The per-class baseline trivially adheres to the conditioning, as each class gets to have its own network. However, when a single network is trained to generate all classes, naive concatenation is unable to successfully inject class information, for either network and for either type of concatenation. For the \textbf{EncoderDecoder} network, basketballs, oranges, cupcakes, pineapples, and fried chicken are all confused with each other. For the \textbf{SkinnyResNet} network, oranges are generated instead of basketballs, and pineapples and fried chicken drumsticks are confused. As seen in Table~\ref{fig:acc_vs_real}, classification accuracy is slightly higher when concatenating all layers ($64.5\%$) versus only the input layer ($62.6\%$), but is low for both.


\vspace{2mm} \noindent \textbf{Does gating effectively inject conditioning?} Using the proposed soft-gating, on the other hand, leads to successful generations. We test variants of soft-gating on the \textbf{SkinnyResNet}, and accuracy is dramatically improved, between $89.6\%$ to $99.6\%$, comparable to using a single generator per class ($97.0\%$).
Among the gating mechanisms, we find that channel-wise multiplication
generates the most realistic images, achieving an AMT fooling rate of $23.4\%$. Interestingly, the fooling rate is higher than the per-class generator of $17.7\%$. Qualitatively, we notice that per-class generators sometimes exhibits artifacts in the background, as seen in the generation of ``moon". We hypothesize with the correct conditioning mechanism, the single generator across multiple classes has the benefit of seeing more training data and finding common elements across classes, such as clean, white backgrounds.


\vspace{2mm} \noindent \textbf{Is gating effective across architectures?} 
As seen in Table~\ref{fig:acc_vs_real}, using channelwise gating instead of naive concatenation improves performance both accuracy and realism \textit{across} architectures. For example, for the \textbf{EncoderDecoder} architecture, gating enables successful generation of the pineapple.
Both quantitatively and qualitatively, results are better for our proposed \textbf{SkinnyResNet} architecture.


\vspace{2mm} \noindent \textbf{Do the generations generalize to unusual outlines?} The training images consist of the outlines corresponding to the geometry of each class. However, an interesting test scenario is whether the technique generalizes to unseen shape and class combinations. In \figref{fig:teaser}, we show that an input circle not only produces circular objects, such as a basketball, watermelon, and cookie, but also noncircular objects such as strawberry, pineapple, and cupcake. Note that both the pineapple crown and bottom are generated, even without any structural indication of these parts in the outline.

\section{Discussion}


We present a two-stage approach for interactive object generation, centered around the idea of a shape completion intermediary. This step both makes training more stable and also allows us to give coarse geometric feedback to the user, which they can choose to integrate as they desire.

\section*{Acknowledgements}
\noindent AG, PKD, and PHST are supported by the ERC grant ERC-2012-AdG, EPSRC grant Seebibyte EP/M013774/1, EPSRC/MURI grant EP/N019474/1 and would also like to acknowledge the Royal Academy of Engineering and FiveAI. Part of the work was done while AG was an intern at Adobe.


{\small
\bibliographystyle{ieee_fullname}
\bibliography{gatedblocks}
}

\newpage

\begin{figure*}[t]
    \centering
    \includegraphics[width=\linewidth,trim={2.6cm 0 1.8cm 0},clip]{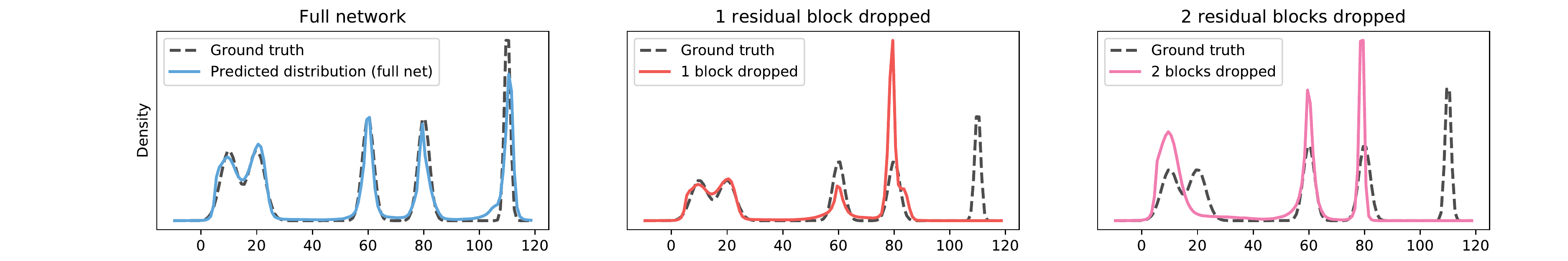}
    \caption{{\bf 1D Mixture of Gaussians.} {\bf (Left)} Samples from a residual network (blue-dotted) closely approximate the training distribution (black). {\bf (Mid)} Removing one residual block removes one mode of the predicted distribution. {\bf (Right)} Removing two blocks drops two modes. Note that samples stay mostly ``on-manifold" of the ground truth distribution.
    }\label{fig:onedexperiment}
\end{figure*}

\section{Insights on Gating Mechanism}
We demonstrate the intuition behind our gating mechanism with a toy experiment where a generative network models a 1D mixture of Gaussians, comprised of five components~\figref{fig:onedexperiment}. 
For this test, the generator and discriminator architectures consist of only residual blocks, where each residual block is composed of fully connected layers. 
%
The generator is conditioned on a latent vector $z$ and is trained to approximate the distribution, as seen in Fig.~\ref{fig:onedexperiment} (left). 
Removing a single residual block, in the spirit of~\cite{veit2016residual}, leads to the disappearance of a mode from the predicted distribution. 
Removal of another block leads to further removal of another mode, as seen in Fig.~\ref{fig:onedexperiment}  (mid, right). 
This experiment suggests that residual blocks arrange themselves  naturally into modeling parts of a distribution, which motivates our use of a gating network where the network learns which blocks (or alternatively, which channels) to attend to for each object class. 

\subsection{Network Architecture}
The architecture was designed to reproduce some of the experiments performed by \cite{veit2016residual} by removing blocks and observing the resulting generated distribution.
While our network is deeper (16 layers of residual blocks) than required for similar experiments e.g., in MAD-GAN \cite{ghosh2017multi}, Mode Regularized GAN \cite{che2016mode} and Unrolled GAN \cite{metz2017unrolledGAN}, we use only 4 neurons in each residual block of the generator and discriminator (Tables~\ref{table:1d_G}~\&~\ref{table:1d_D}) compared to fully connected versions in which there consisted of connections between 256 neurons in the preceding layer to 256 neurons in the current layer. Thus although the number of parameters is much lower, the network learns the distribution quite accurately. The architecture used in this experiment inspired the design of the skinny Resnet architecture as described later.



\begin{table}[h]
\caption{\textbf{ResBlock}} 
\centering 
\begin{tabular}{c} 
\toprule
\textbf{F(x)}\\\midrule
Linear\\ 
ReLU() \\
Linear\\
\bottomrule 
\end{tabular}
\label{table:resblock} 
\end{table}

\begin{table}[h]
\caption{\textbf{Generator for 1D setting}} 
\centering 
\begin{tabular}{l c c}
\toprule
\textbf{Layer} & \textbf{Neurons} & \textbf{Num Layers} \\ \midrule
Linear & 10 $\rightarrow$ 4 & 1  \\ 
ResBlock & 4 & 16 \\ 
Linear & 4$\rightarrow$ 1 & 1 \\ 
\bottomrule 
\end{tabular}
\label{table:1d_G} 
\end{table}

\begin{table}[ht]
\caption{\textbf{Discriminator for 1D setting}} 
\centering 
\begin{tabular}{l c c}
\toprule
\textbf{Layer} & \textbf{Neurons} & \textbf{Num Layers} \\ \midrule
Linear & 1 $\rightarrow$ 4 & 1  \\ 
ResBlock & 4 & 16 \\ 
Linear & 4 $\rightarrow$ 1 & 1 \\
Sigmoid & 1 & 1 \\
\bottomrule 
\end{tabular}
\label{table:1d_D} 
\end{table}

\section{Shape Completion Details}


For shape completion, training and testing inputs were created using by placing occluders of 3 sizes (64$\times$64, 128$\times$128, 192$\times$192) on top of full sketches or outlines.
For each size, 25 partial sketches/outlines were created by random placement of the occluder, thus leading to 75 partial versions to be completed from a single sketch/outline.

\begin{table}[h]
\caption{\textbf{Shape Generator}} 
\centering 
\begin{tabular}{l c c}
\toprule
\textbf{Layer} & \textbf{Output Size} & \textbf{Filter} \\ \midrule
\small{Fully Connected} & \small{$512 \times 4 \times 4$} & \small{$256 \rightarrow 512 \times 4 \times 4$} \\
\small{Reshape} & \small{$512 \times 4 \times 4$} & - \\
\midrule
\small{Resnet-Block} & \small{$512 \times 4 \times 4$} & \small{$512 \rightarrow 512 \rightarrow 512$} \\
\small{Resnet-Block} & \small{$512 \times 4 \times 4$} & \small{$512 \rightarrow 512 \rightarrow 512$} \\
\small{Sparse-Resnet-Block} & \small{$512 \times 4 \times 4$} & \small{$1 \rightarrow 512 \rightarrow 512$} \\
\small{NN-Upsampling} & \small{$512 \times 8 \times 8$} & - \\
\midrule
\small{Resnet-Block} & \small{$512 \times 8 \times 8$} & \small{$512 \rightarrow 512 \rightarrow 512$} \\
\small{Resnet-Block} & \small{$512 \times 8 \times 8$} & \small{$512 \rightarrow 512 \rightarrow 512$} \\
\small{Sparse-Resnet-Block} & \small{$512 \times 8 \times 8$} & \small{$1 \rightarrow 512 \rightarrow 512$} \\
\small{NN-Upsampling} & \small{$512 \times 16 \times 16$} & - \\
\midrule
\small{Resnet-Block} & \small{$256 \times 16 \times 16$} & \small{$512 \rightarrow 256 \rightarrow 256$} \\
\small{Resnet-Block} & \small{$256 \times 16 \times 16$} & \small{$256 \rightarrow 256 \rightarrow 256$} \\
\small{Sparse-Resnet-Block} & \small{$256 \times 16 \times 16$} & \small{$1 \rightarrow 256 \rightarrow 256$} \\
\small{NN-Upsampling} & \small{$256 \times 32 \times 32$} & - \\
\midrule
\small{Resnet-Block} & \small{$128 \times 32 \times 32$} & \small{$256 \rightarrow 128 \rightarrow 128$} \\
\small{Resnet-Block} & \small{$128 \times 32 \times 32$} & \small{$128 \rightarrow 128 \rightarrow 128$} \\
\small{Sparse-Resnet-Block} & \small{$128 \times 32 \times 32$} & \small{$128 \rightarrow 128 \rightarrow 128$} \\
\small{NN-Upsampling} & \small{$128 \times 64 \times 64$} & - \\
\midrule
\small{Resnet-Block} & \small{$64 \times 64 \times 64$} & \small{$128 \rightarrow 64 \rightarrow 64$} \\
\small{Resnet-Block} & \small{$64 \times 64 \times 64$} & \small{$64 \rightarrow 64 \rightarrow 64$} \\
\small{Sparse-Resnet-Block} & \small{$64 \times 64 \times 64$} & \small{$1 \rightarrow 64 \rightarrow 64$} \\
\small{NN-Upsampling} & \small{$64 \times 128 \times 128$} & - \\
\midrule
\small{Resnet-Block} & \small{$32 \times 128 \times 128$} & \small{$64 \rightarrow 32 \rightarrow 32$} \\
\small{Resnet-Block} & \small{$32 \times 128 \times 128$} & \small{$32 \rightarrow 32 \rightarrow 32$} \\
\small{Conv2D} & \small{$1 \times 128 \times 128$} & \small{$ 64 \rightarrow 1$} \\
\bottomrule 
\end{tabular}
\label{table:G_S} 
\end{table}

\begin{table}[h]
\caption{\textbf{Shape Discriminator}} 
\centering 
\begin{tabular}{l c c}
\toprule
\textbf{Layer} & \textbf{Output Size} & \textbf{Filter} \\ \midrule
\small{Conv2D} & \small{$32 \times 128 \times 128$} & \small{$2 \rightarrow 32$} \\
\midrule
\small{Resnet-Block} & \small{$32 \times 128 \times 128$} & \small{$32 \rightarrow 32 \rightarrow 32$} \\
\small{Resnet-Block} & \small{$64 \times 128 \times 128$} & \small{$32 \rightarrow 32 \rightarrow 64$} \\
\small{Sparse-Resnet-Block} & \small{$64 \times 128 \times 128$} & \small{$2 \rightarrow 128 \rightarrow 128$} \\
\small{Avg-Pool2D} & \small{$64 \times 64 \times 64$} & - \\
\midrule
\small{Resnet-Block} & \small{$64 \times 64 \times 64$} & \small{$64 \rightarrow 64 \rightarrow 64$} \\
\small{Resnet-Block} & \small{$128 \times 64 \times 64$} & \small{$64 \rightarrow 64 \rightarrow 128$} \\
\small{Sparse-Resnet-Block} & \small{$128 \times 64 \times 64$} & \small{$2 \rightarrow 64 \rightarrow 64$} \\
\small{Avg-Pool2D} & \small{$128 \times 32 \times 32$} & - \\
\midrule
\small{Resnet-Block} & \small{$128 \times 32 \times 32$} & \small{$128 \rightarrow 128 \rightarrow 128$} \\
\small{Resnet-Block} & \small{$256 \times 32 \times 32$} & \small{$128 \rightarrow 128 \rightarrow 256$} \\
\small{Sparse-Resnet-Block} & \small{$256 \times 32 \times 32$} & \small{$2 \rightarrow 256 \rightarrow 256$} \\
\small{Avg-Pool2D} & \small{$256 \times 16 \times 16$} & - \\
\midrule
\small{Resnet-Block} & \small{$256 \times 16 \times 16$} & \small{$256 \rightarrow 256 \rightarrow 256$} \\
\small{Resnet-Block} & \small{$512 \times 16 \times 16$} & \small{$256 \rightarrow 256 \rightarrow 512$} \\
\small{Sparse-Resnet-Block} & \small{$512 \times 16 \times 16$} & \small{$2 \rightarrow 512 \rightarrow 512$} \\
\small{Avg-Pool2D} & \small{$512 \times 8 \times 8$} & - \\
\midrule
\small{Resnet-Block} & \small{$512 \times 8 \times 8$} & \small{$512 \rightarrow 512 \rightarrow 512$} \\
\small{Resnet-Block} & \small{$512 \times 8 \times 8$} & \small{$512 \rightarrow 512 \rightarrow 512$} \\
\small{Sparse-Resnet-Block} & \small{$512 \times 8 \times 8$} & \small{$2 \rightarrow 512 \rightarrow 512$} \\
\small{Avg-Pool2D } & \small{$512 \times 4 \times 4$} & - \\
\midrule
\small{Resnet-Block} & \small{$512 \times 4 \times 4$} & \small{$512 \rightarrow 512 \rightarrow 512$} \\
\small{Resnet-Block} & \small{$512 \times 4 \times 4$} & \small{$512 \rightarrow 512 \rightarrow 512$} \\
\small{Fully Connected} & \small{$N_{classes}$} & \small{\small{$ 512 \cdot 4 \cdot 4 \rightarrow N_{classes}$}} \\
\bottomrule 
\end{tabular}
\label{table:D_S} 
\end{table}

The generator architecture for the shape completion is depicted in table \tabref{table:G_S} while the discriminator architecture is depicted in table \tabref{table:D_S}. 
The architecture is almost the same as \cite{mescheder2018training} except for the sparse Resnet blocks used for injecting conditioning via multiple scales. 
The sparse Resnet blocks first resize the input conditioning (for example, the partial user strokes), and then convert the feature map into the correct number of channels using a Resnet block to add to the feature activation.
This occurs just prior to the upsampling step in the generator and just prior to the avg pool step in the discriminator. 



\begin{figure*}[t]
\centering
\begin{tabular}{*{3}{c@{\hspace{3px}}}}
\textbf{Blockwise Gating} & \textbf{Channelwise Gating} \vspace{-1mm} & \\
\includegraphics[height=3.35cm,trim={.15cm 0 0 .4cm}, clip]{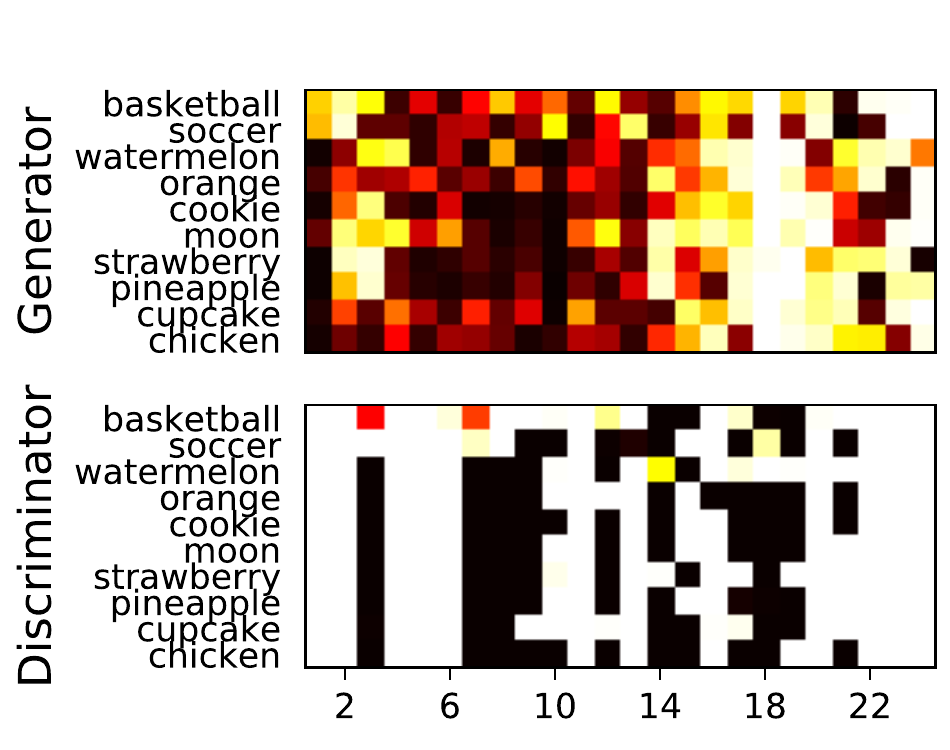} &
\includegraphics[height=3.35cm,trim={0 0 0 .4cm}, clip]{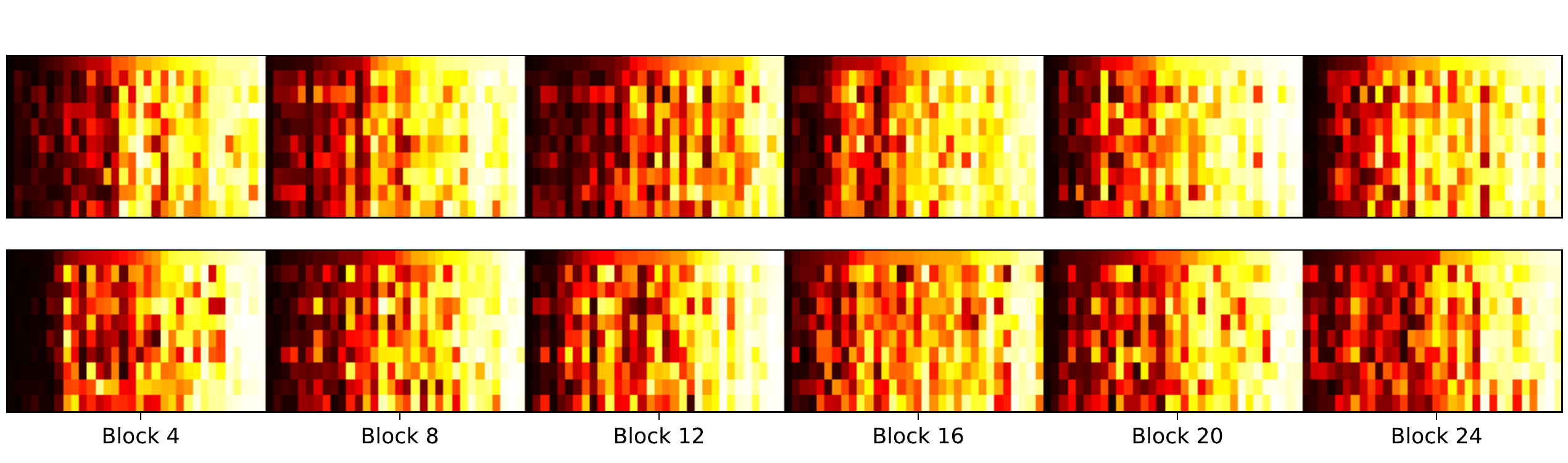} & 
\includegraphics[height=3.35cm,trim={5.8cm 0 .2cm .4cm}, clip]{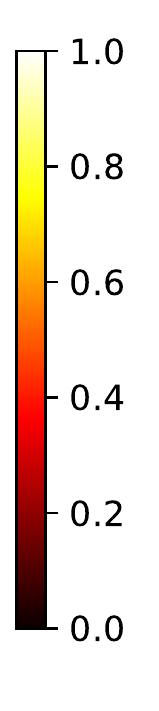}
\\
\end{tabular}
\vspace{-2mm}
\caption{\label{fig:alpha_heat}
\textbf{Learned gating parameters.} We show the soft-gating parameters for {\bf (left)} blockwise and {\bf (right)} channelwise gating for the {\bf (top)} generator and {\bf (bot)} discriminator. Black indicates
completely off, and white indicates
completely on. For channelwise, a subset (every 4th) of blocks is shown. Within each block, channels are sorted in ascending order of the first category. The nonuniformity of each columns indicates that different channels are used more heavily for different classes.
\vspace{-3mm}
}
\vspace{-2mm}
\end{figure*}

\section{Outline$\rightarrow$Image Network Architecture}
The network architecture is based on our observations that deeper, narrower networks perform better when capturing multi-modal data distributions. 
The second guiding principle in the design of the architecture is that the different blocks should have similar number of channels so that the gating hypernetwork can distribute the modes between the blocks efficiently. 
Finally, we apply gating to the residual blocks responsible for upsampling and downsampling as well, in order to allow for better fine-grained control on the generation process. 
\tabref{table:convresblock} shows the Convolution Residual Block which does not change the spatial resolution of the activation volume, 
\tabref{table:downconvresblock} shows the Downsampling Residual Block which reduces the activation volume to half the spatial resolution, 
\tabref{table:upconvresblock} shows the Upsampling Residual Block which increases the activation volume to twice the spatial resolution.
In the case of gating (either block wise/channel-wise) the gating is applied on the $F(x)$ of each network. 
The shortcut branch represented in \tabref{table:upconvresblock} and \tabref{table:downconvresblock} represents the branch of the Resnet which is added to $F(x)$ branch. In these scenarios since the resolution of $x$ changes in $F(x)$, the shortcut also has a similar upsampling/downsampling layer.

\begin{table}[ht]
    \makegapedcells
        \centering 
        \begin{tabular}{c} 
        \toprule
        \textbf{F(x)}\\
        \midrule
        Conv2d \\
        InstanceNorm\\ 
        ReLU() \\
        Conv2d \\
        InstanceNorm\\ 
        ReLU() \\
        \bottomrule 
        \end{tabular}
        \caption{\textbf{ConvResblock}} 
        \label{table:convresblock} 
\end{table}

\begin{table}
        \centering 
        \begin{tabular}{c} 
        \toprule 
        \textbf{F(x)}\\
        \midrule 
        Upsample (Nearest Neighbor) \\
        ReflectionPad \\
        Conv2d \\
        InstanceNorm\\ 
        ReLU() \\
        Conv2d \\
        InstanceNorm\\ 
        ReLU() \\
        \midrule 
        \textbf{Shortcut Branch}\\
        \midrule 
        Upsample (Bilinear) \\
        ReflectionPad\\
        Conv2d \\
        \bottomrule 
        \end{tabular}
        \caption{        \label{table:upconvresblock} \textbf{UpConvResblock}} 
\end{table}

\begin{table}
        \centering 
        \begin{tabular}{c} 
        \toprule 
        \textbf{F(x)}\\
        \midrule
        Avgpool 2d \\
        ReflectionPad \\
        Conv2d \\
        InstanceNorm\\ 
        ReLU() \\
        Conv2d \\
        InstanceNorm\\ 
        ReLU() \\
        \midrule
        \textbf{Shortcut Branch}\\
        \midrule 
        Avgpool 2d \\
        ReflectionPad\\
        Conv2d \\
        \bottomrule
        \end{tabular}
        \caption{\label{table:downconvresblock} \textbf{DownConvResblock}} 
\end{table}

\begin{table}[ht]
\caption{\textbf{Gated Resnet G: Scribble Dataset}}
\centering 
\begin{tabular}{l c c} 
\toprule
\textbf{Layer} & \textbf{Filter} & \textbf{Num Layers} \\
\midrule
Conv2d & 3 $\rightarrow$ 32 & 1\\
InstanceNorm & 32 & 1 \\ 
ReLU() & 32 & 1\\
\hdashline
\textbf{Gated}-ConvResBlock & 32 & 3\\
\textbf{Gated}-DownConvResBlock & 32 & 3\\
\textbf{Gated}-ConvResBlock & 32 & 12\\
\textbf{Gated}-UpConvResBlock & 32 & 3\\
\textbf{Gated}-ConvResBlock & 32 & 3\\
\hdashline
Conv2d & 32$\rightarrow$3 & 1 \\
Tanh() & 3 & 1 \\
\bottomrule
\end{tabular}
\label{table:resnet_g_scribble} 
\end{table}

\begin{table}[ht]
\caption{\textbf{Gated Resnet D: Scribble Dataset}}
\centering 
\begin{tabular}{l c c} 
\toprule
\textbf{Layer} & \textbf{Filter} & \textbf{Num Layers} \\
\midrule
Conv2d & 6 $\rightarrow$ 32 & 1\\
\hdashline
\textbf{Gated}-ConvResBlock & 32 & 3\\
\textbf{Gated}-DownConvResBlock & 32 & 4\\
\textbf{Gated}-ConvResBlock & 32 & 17\\
\hdashline
Conv2d & 32$\rightarrow$1 & 1 \\
Sigmoid() & 1 & 1 \\
\bottomrule
\end{tabular}
\label{table:resnet_d_scribble} 
\end{table}

\begin{table}[ht]
\caption{\textbf{Gating Hypernetwork} $dim^{gate}$ is the number of blocks in the case of Block Wise Gating and the number of channels in the case of Channel Wise Gating. In case of affine its twice of each since the $\beta$ is of the same dimension }
\centering 
\begin{tabular}{l c c} 
\toprule
\textbf{Layer} & \textbf{Filter/Shape} & \textbf{Num Layers} \\
\midrule
Embedding & $dim^{embed}$ & 1 \\
Conv1d & 1 $\rightarrow$ 16 & 1\\
ResBlock1D & 16 & 16\\
Reshape & 16$\times dim^{embed}$ & 1\\
Linear & 16$\times dim^{embed} \rightarrow dim^{gate} $& 1\\
\bottomrule
\end{tabular}
\label{table:resnet_gating} 
\end{table}



\begin{figure}
    \centering
    \includegraphics{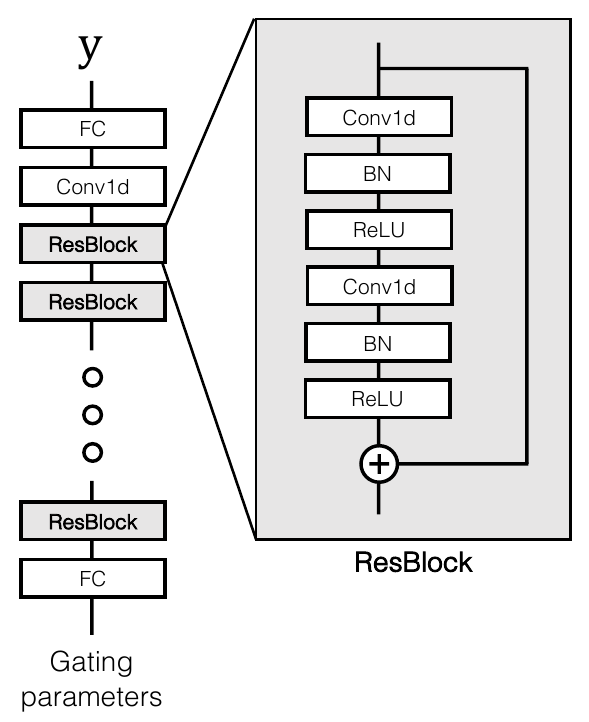}
    \caption{Gating Hypernetwork architecture.}
    \label{fig:gating_arch}
\end{figure}

\subsection{Gating Hypernetwork}
The gating hypernetwork was also designed using Resnet blocks. We use 1D convolutions in the Resnet block \tabref{table:resblock1D} to reduce the number of parameters and use BatchNormalization to speed up the training of the network responsible for prediction gating parameters. 
Class conditioning is first passed through an embedding layer to obtain a representation of the class which is further processed by the Resnet blocks. 
The same network is used for the various forms of gating. In case of block wise gating, the number of outputs $dim^{gate}$ for this network is equal to the number of blocks used in the main network.
In the case of an affine transformation, the network predicts an equal number of biases for each the block. 
In the case of channel-wise gating, the number of predicted parameters $dim^{gate}$ is equal to $num^{channels}\times num^{blocks}$ since each residual block consists of equal number of channels.
$\alpha$ was constrained between 0 and 1 corresponding to selecting or rejecting a block, while the $\beta$ was restricted between -1 and 1 when used. 
In the original AdaIN case, parameters are unrestricted, but we found we had to constrain parameters between -1 and 1 in order for the network to perform well. 

\begin{table}[ht]
\caption{\textbf{ResBlock1D}} 
\centering 
\begin{tabular}{c} 
\toprule
\textbf{F(x)}\\\midrule
Conv1D\\ 
BatchNorm1D\\
ReLU \\
Conv1D\\
BatchNorm1D\\
ReLU \\
\bottomrule 
\end{tabular}
\label{table:resblock1D} 
\end{table}

\section{Distribution of Alphas}
\begin{figure*}[t]
    \centering
    \includegraphics[width=\linewidth]{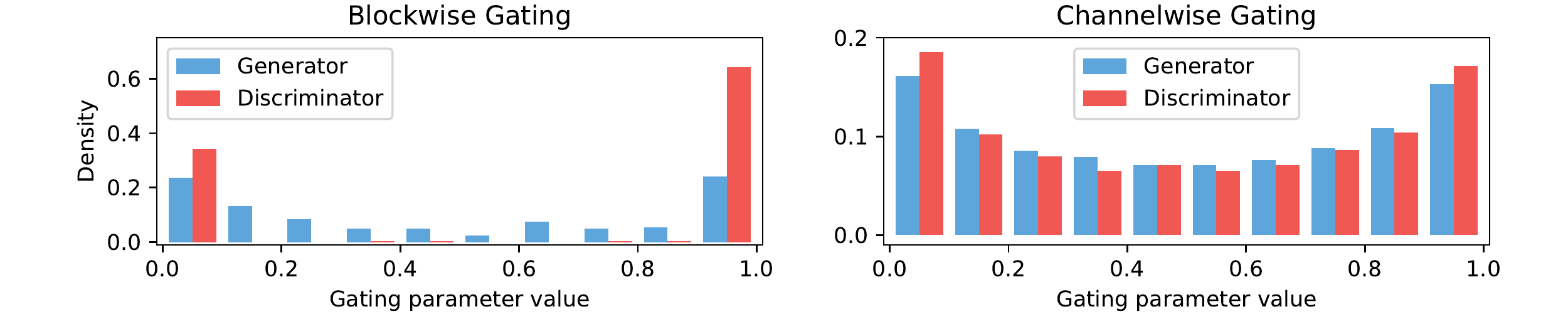}
    \caption{We show the distribution of the $\alpha$ values. Typically, they are closer to the extremes (0 or 1) rather than the intermediate values.}
    \label{fig:alpha_hist}
\end{figure*}

A histogram of the distribution of the various alphas for the block-wise setting and the channel-wise setting is shown in \figref{fig:alpha_hist}. Even without an explicit sparsity constraint, the alphas are pushed near the extremes. 

\begin{figure*}[t]
\centering
\begin{tabular}{*{2}{c@{\hspace{3px}}}}
\includegraphics[height=3.15cm,trim={6.0cm 0 7.6cm .4cm}, clip]{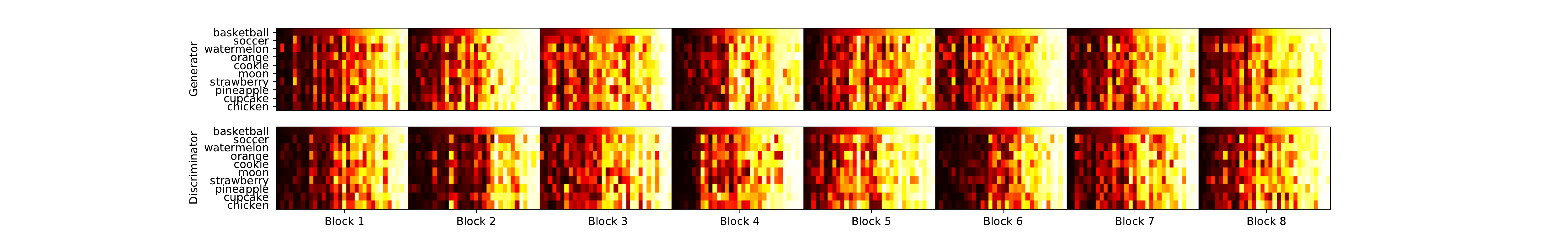} & 
\includegraphics[height=3.15cm,trim={5.8cm 0 .2cm .4cm}, clip]{paper_images/alpha_legend.pdf}
\\
\includegraphics[height=3.15cm,trim={6.0cm 0 7.6cm .4cm}, clip]{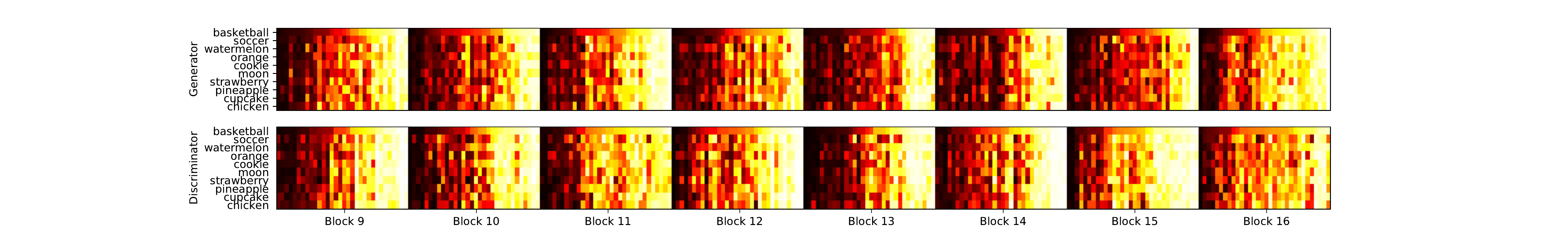} & 
\includegraphics[height=3.15cm,trim={5.8cm 0 .2cm .4cm}, clip]{paper_images/alpha_legend.pdf}
\\
\includegraphics[height=3.15cm,trim={6.0cm 0 7.6cm .4cm}, clip]{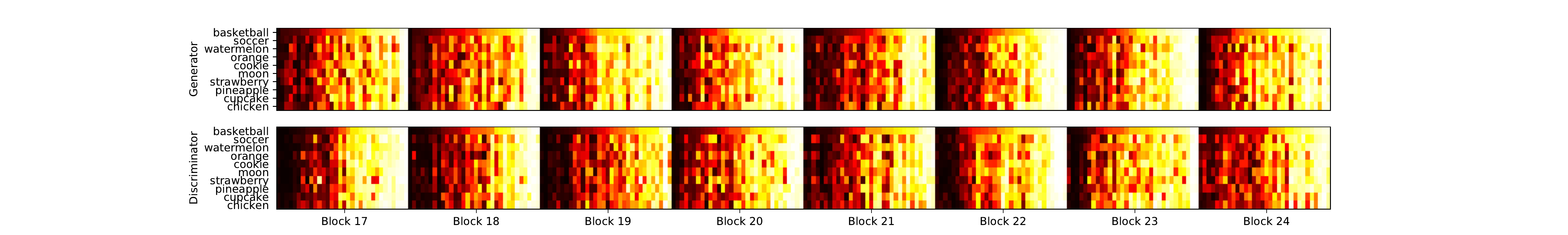} & 
\includegraphics[height=3.15cm,trim={5.8cm 0 .2cm .4cm}, clip]{paper_images/alpha_legend.pdf}
\\

\end{tabular}
\caption{\label{fig:alpha_heat}
\textbf{Learned channel-wise gating parameters.} We show the soft-gating parameters for channelwise gating for the {\bf (top)} generator and {\bf (bot)} discriminator. Black indicates
completely off, and white indicates
completely on. We show all 24 blocks. The nonuniformity of each columns indicates that different channels are used more heavily for different classes.
}
\end{figure*}

\section{Unusual Shapes for Various Classes}

\begin{figure*}[t]
    \centering
    \includegraphics[width=\linewidth,trim={0 0 4.5cm 0},clip]{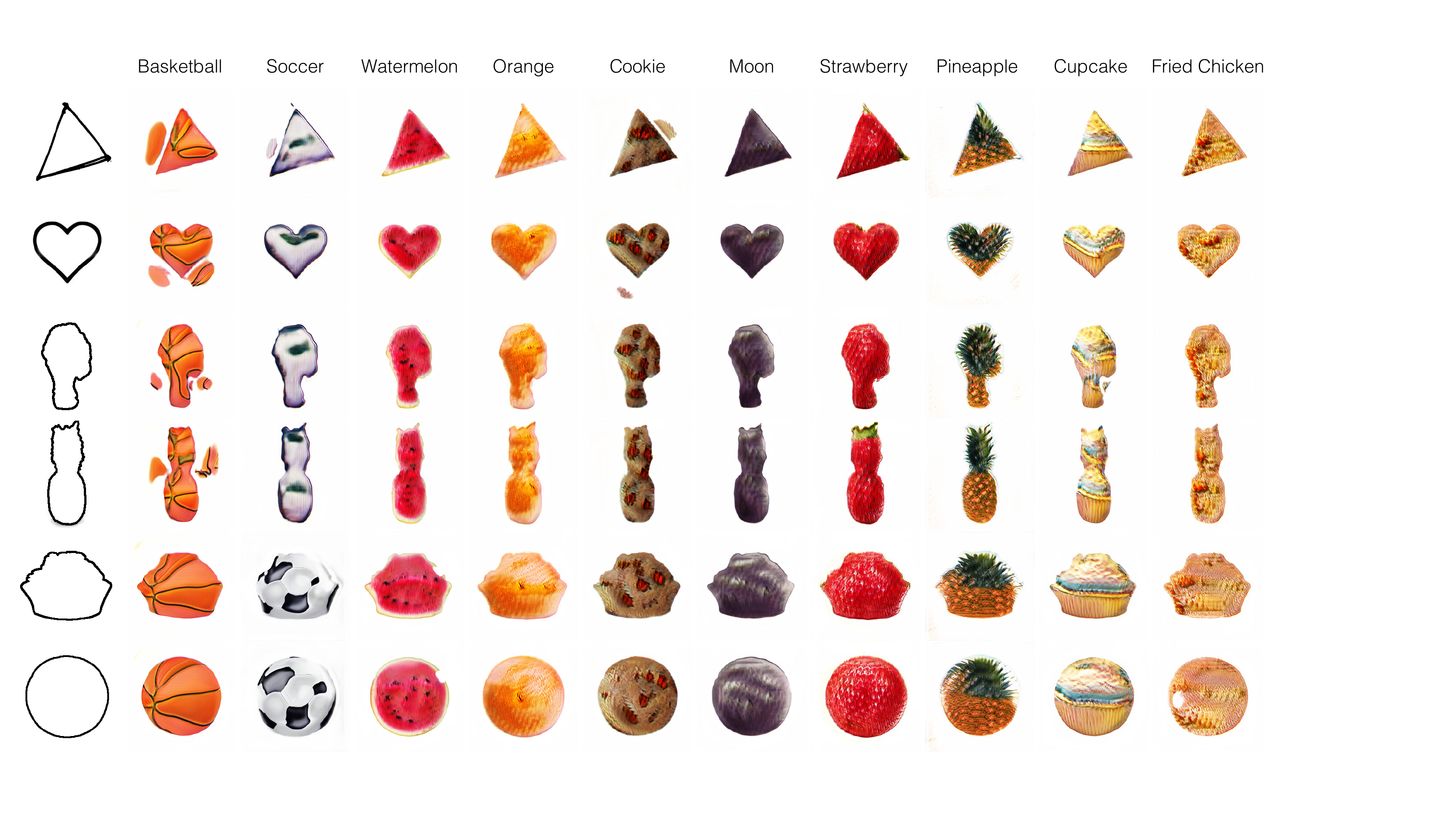}
    \caption{{\bf Channel-Wise Gating:} We observe that the technique can also generate images with shapes not seen by that class during training time.}
    \label{fig:channel_shapes}
\end{figure*}

As evident from \figref{fig:channel_shapes} 
the gated generative techniques extend to shapes it never was shown while training.

\end{document}


\title{Interactive Sketch \& Fill: Multiclass Sketch-to-Image Translation \\ Supplementary Material}


\author{
Arnab Ghosh$^{1}$ \hspace{8mm} Richard Zhang$^{2}$ \hspace{8mm} Puneet K. Dokania$^{1}$ \\
Oliver Wang$^{2}$ \hspace{8mm} Alexei A. Efros$^{2,3}$ \hspace{8mm} Philip H. S. Torr$^{1}$ \hspace{8mm} Eli Shechtman$^{2}$ \\
\\
$^{1}$University of Oxford \hspace{15mm} $^{2}$Adobe Research \hspace{15mm} $^{3}$UC Berkeley \\
}

\maketitle

\section{User Study (User Interface):}
In order to analyze how users perceive the effects of different components of the user interface, we conducted a pilot study with 14 participants who were all graduate students of computer science in the age group of 20-30 and about 30\% of them women and 70\% men with little to no artistic experience. 
We attempted to study three main variables: \textbf{outlines vs edges, interactivity vs no interactivity, and completion hints vs no completion hints.} 
Each participant was instructed to use each variant of our system, presented in a random order, in order to generate realistic images from 5 random classes. 
Each user was asked to rate his/her performance on the NASA Task Load Index (lower the better). \cite{hart1988development}.

\begin{figure*}[t]
    \centering
    \includegraphics[width=\linewidth]{user_study/barplots.pdf}
    \caption{{\bf NASA TLX Score Comparisons (lower is better): }
    The NASA TLX Questionnaire consists of 6 different subjective scales the raw scores for the various interfaces are depicted in the figure, the aggregate is computed using the weight of each subjective subscale using a pair-wise importance ranking performed by the subject. We see that our full system systematically obtains the lowest scores in all the subjective subscales while the one with the dense edge inputs obtains the highest.\label{fig:nasa-tlx}
    \vspace{-2mm}
    }
    \vspace{-2mm}
\end{figure*}

\begin{table*}[h]
\resizebox{\linewidth}{!}{
\caption{\textbf{User Study: User Interfaces}} 
\centering 
\begin{tabular}{l c c c c}
\toprule
\textbf{Interface} & \textbf{Input} & \textbf{Recommendation} &  \textbf{Interactive} & \textbf{Mean Aggregate Score} \\ \midrule
Non Interactive & Outline & \checkmark &  & 29.01\\ 
Dense Edge Inputs & Edge & \checkmark & \checkmark & 47.50 \\ 
No outline completion & Outline & & \checkmark & 28.29\\ 
\textbf{Our Full System} & \textbf{Outline} & \textbf{\checkmark} & \textbf{\checkmark} & \textbf{20.83} \\
\bottomrule 
\end{tabular}
\label{table:user_study} 
}
\end{table*}


\subsection{NASA TLX: Details}
NASA TLX questionnaire \cite{hart1988development} is a standard questionnaire used by psychological studies which gauges workload of a task on 6 metrics namely: Mental Demand, Physical Demand, Temporal Demand, Performance, Effort and Frustration. Since different people interpret these terms differently first they are provided with a pairwise comparison of each of the terms being the most important contributor to the workload for the given task. The ratings on each of these scales are weighted by the relative importance (obtained from the pairwise comparisons) and averaged out to get a score from 0-100 which represents the workload for the current task (lower is better).

\tabref{table:user_study} shows the mean aggregate scores obtained for each interface, aggregate scores are obtained by weighting each subjective score on the individual metric by the user's pairwise importance ranking. For a more detailed analysis into the individual metrics please refer to \figref{fig:nasa-tlx} which provides the raw scores (unweighted) for each individual metric.


As depicted in \tabref{table:user_study} and \figref{fig:nasa-tlx} we can see that:

\begin{itemize}
    \item \textbf{Outlines vs Edges:} Mean score for Our Full System is lesser than Dense Edge Inputs thus indicating that users prefer a model trained with outlines compared to one trained with edges.
    \item \textbf{Interactivity vs No Interactivity:} Mean score for No outline completion and Our Full System which are outline based but has interactivity shows lower scores than Non Interactive which doesn't have interactivity. This shows that users prefer interactivity to non-interactivity.
    \item \textbf{Sketch Recommendation vs No Sketch Recommendation:} Mean score for Our Full System which has outline-completion based recommendation for the sketch are lower than for No outline completion thus indicating that users prefer recommended outlines over no outline recommendation.
\end{itemize}

A further larger scale study might be needed to fully validate the hypotheses. However, this pilot study shows that a recommendation system (sketch completion) where the outline completion is shown along along with the corresponding generated image, was given the highest preference by the users.

\section{Shape Completion: Details}

The Bicycle GAN model \cite{zhu2017toward} was used to train the shape completion model with minor modifications. In order for the BicycleGAN model to generate the shape completion appropriate for the correct class the class condition was concatenated to the input as a one hot encoding in the Generator, Discriminator and the Encoder. Since the class condition worked decently we didn't try introducing gating in those scenarios but it can potentially be introduced although the gating might potentially produce multimodal results. The input domain images for the model consisted of the partial versions of the sketches or outlines and the target domain for the model consisted of the completed sketches or outlines. 

The training and testing input partial sketches and partial outlines were created using the same modus operandi whereby for each sketch/outline occluders of 3 sizes (64x64,128x128,192x192) were used and for each size of occluder 25 partial sketches/outlines were created, thus creating 75 partial versions from a single sketch/outline.

\section{Appearance Generation: Details}
\begin{figure*}[t]
    \centering
    \includegraphics[width=\linewidth]{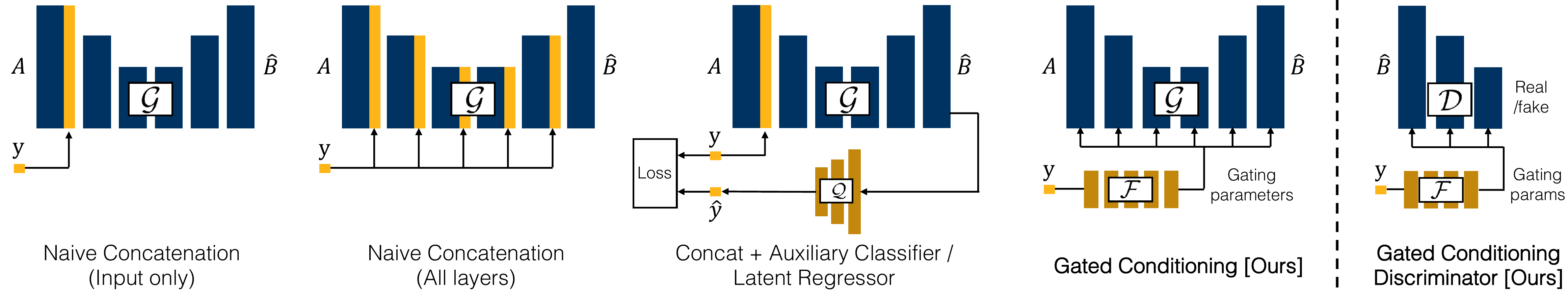}
    \caption{{\bf Conditioning injection variants.}
    The conditioner can be naively incorporated in a generator through simple concatenation in {\bf (left)} the input layer only or {\bf (mid-left)} in all layers. {\bf (mid)} The network can be further encouraged to use the conditioning through a learned network, using either a classification objective for categorical conditioning~\cite{odena2016conditional,chen2016infogan} or a regression objective for continuous conditioning. {\bf (Mid-right)} We train a network on the conditioner to predict parameters which guide softly-gated units in the main network. {\bf (Right)} These conditioning options can be correspondingly applied to a discriminator. We propose using soft-gating on the discriminator as well.\label{fig:arch-inj}
    \vspace{-2mm}
    }
    \vspace{-2mm}
\end{figure*}

For more details about the various architectures described in Section 4.2 in the main paper we show the schematic representation of all the above mentioned models can be observed in \figref{fig:arch-inj} and Figure 5 in the main paper.

\begin{figure*}[t]
\centering
\begin{tabular}{*{3}{c@{\hspace{3px}}}}
\textbf{Blockwise Gating} & \textbf{Channelwise Gating} \vspace{-1mm} & \\
\includegraphics[height=3.35cm,trim={.15cm 0 0 .4cm}, clip]{paper_images/alphas_block.pdf} &
\includegraphics[height=3.35cm,trim={0 0 0 .4cm}, clip]{paper_images/alphas_chan.pdf} & 
\includegraphics[height=3.35cm,trim={5.8cm 0 .2cm .4cm}, clip]{paper_images/alpha_legend.pdf}
\\
\end{tabular}
\vspace{-2mm}
\caption{\label{fig:alpha_heat}
\textbf{Learned gating parameters.} We show the soft-gating parameters for {\bf (left)} blockwise and {\bf (right)} channelwise gating for the {\bf (top)} generator and {\bf (bot)} discriminator. Black indicates
completely off, and white indicates
completely on. For channelwise, a subset (every 4th) of blocks is shown. Within each block, channels are sorted in ascending order of the first category. The nonuniformity of each columns indicates that different channels are used more heavily for different classes.
\vspace{-3mm}
}
\vspace{-2mm}
\end{figure*}

\section{1D Network Architecture}
The architecture was designed hoping to reproduce some of the experiments performed by \cite{veit2016residual} by removing blocks and observing the resulting generated distribution, hence a design choice of having a deep network but keeping the neurons in each block lesser in order to reduce chances of overfitting/overparametrizing for a relatively simpler task. The interesting aspect was that although the network was deeper (16 layers of residual blocks) than required for similar experiments in MAD-GAN \cite{ghosh2017multi}, Mode Regularized GAN \cite{che2016mode} and Unrolled GAN \cite{metz2017unrolledGAN}, there were only 4 neurons in each residual block of the generator and discriminator (Tables~\ref{table:1d_G}~\&~\ref{table:1d_D}) compared to fully connected versions in which there consisted of connections between 256 neurons in the preceding layer to 256 neurons in the current layer. Thus although the number of parameters were much less, the network learned the distribution quite accurately. The architecture used in this experiment inspired the design of the skinny Resnet architecture as described later.

\begin{table}[ht]
\caption{\textbf{ResBlock}} 
\centering 
\begin{tabular}{c} 
\toprule
\textbf{F(x)}\\\midrule
Linear\\ 
ReLU() \\
Linear\\
\bottomrule 
\end{tabular}
\label{table:resblock} 
\end{table}

\begin{table}[ht]
\caption{\textbf{Generator for 1D setting}} 
\centering 
\begin{tabular}{l c c}
\toprule
\textbf{Layer} & \textbf{Neurons} & \textbf{Num Layers} \\ \midrule
Linear & 10 $\rightarrow$ 4 & 1  \\ 
ResBlock & 4 & 16 \\ 
Linear & 4$\rightarrow$ 1 & 1 \\ 
\bottomrule 
\end{tabular}
\label{table:1d_G} 
\end{table}



\begin{table}[ht]
\caption{\textbf{Discriminator for 1D setting}} 
\centering 
\begin{tabular}{l c c}
\toprule
\textbf{Layer} & \textbf{Neurons} & \textbf{Num Layers} \\ \midrule
Linear & 1 $\rightarrow$ 4 & 1  \\ 
ResBlock & 4 & 16 \\ 
Linear & 4 $\rightarrow$ 1 & 1 \\
Sigmoid & 1 & 1 \\
\bottomrule 
\end{tabular}
\label{table:1d_D} 
\end{table}

\section{Outline$\rightarrow$Image Network Architecture}
The network architecture is based on our observations that deeper networks perform better capturing multi-modal data distributions. The second guiding principle in the design of the architecture is that the different blocks should have similar number of channels so that the gating hypernetwork can distribute the modes between the blocks efficiently. Finally, the residual blocks responsible for upsampling and downsampling were gated as well in order to allow more blocks to be gated and hence better fine-grained control on the generation process. 
\tabref{table:convresblock} shows the Convolution Residual Block which does not change the spatial resolution of the activation volume, 
\tabref{table:downconvresblock} shows the Downsampling Residual Block which reduces the activation volume to half the spatial resolution, 
\tabref{table:upconvresblock} shows the Upsampling Residual Block which increases the activation volume to twice the spatial resolution,  and in the case of gating (either block wise/channel-wise) the gating is applied on the $F(x)$ of each network. 
The shortcut branch represented in \tabref{table:upconvresblock} and \tabref{table:downconvresblock} represents the branch of the Resnet which is added to $F(x)$ branch. In these scenarios since the resolution of $x$ changes in $F(x)$ branch hence the shortcut also has a similar upsampling/downsampling layer.

\begin{table}[ht]
    \makegapedcells
        \centering 
        \begin{tabular}{c} 
        \toprule
        \textbf{F(x)}\\
        \midrule
        Conv2d \\
        InstanceNorm\\ 
        ReLU() \\
        Conv2d \\
        InstanceNorm\\ 
        ReLU() \\
        \bottomrule 
        \end{tabular}
        \caption{\textbf{ConvResblock}} 
        \label{table:convresblock} 
\end{table}

\begin{table}
        \centering 
        \begin{tabular}{c} 
        \toprule 
        \textbf{F(x)}\\
        \midrule 
        Upsample (Nearest Neighbor) \\
        ReflectionPad \\
        Conv2d \\
        InstanceNorm\\ 
        ReLU() \\
        Conv2d \\
        InstanceNorm\\ 
        ReLU() \\
        \midrule 
        \textbf{Shortcut Branch}\\
        \midrule 
        Upsample (Bilinear) \\
        ReflectionPad\\
        Conv2d \\
        \bottomrule 
        \end{tabular}
        \caption{        \label{table:upconvresblock} \textbf{UpConvResblock}} 
\end{table}

\begin{table}
        \centering 
        \begin{tabular}{c} 
        \toprule 
        \textbf{F(x)}\\
        \midrule
        Avgpool 2d \\
        ReflectionPad \\
        Conv2d \\
        InstanceNorm\\ 
        ReLU() \\
        Conv2d \\
        InstanceNorm\\ 
        ReLU() \\
        \midrule
        \textbf{Shortcut Branch}\\
        \midrule 
        Avgpool 2d \\
        ReflectionPad\\
        Conv2d \\
        \bottomrule
        \end{tabular}
        \caption{\label{table:downconvresblock} \textbf{DownConvResblock}} 
\end{table}

\begin{table}[ht]
\caption{\textbf{Gated Resnet G:Scribble Dataset}}
\centering 
\begin{tabular}{l c c} 
\toprule
\textbf{Layer} & \textbf{Filter} & \textbf{Num Layers} \\
\midrule
Conv2d & 3 $\rightarrow$ 32 & 1\\
InstanceNorm & 32 & 1 \\ 
ReLU() & 32 & 1\\
\hdashline
\textbf{Gated}-ConvResBlock & 32 & 3\\
\textbf{Gated}-DownConvResBlock & 32 & 3\\
\textbf{Gated}-ConvResBlock & 32 & 12\\
\textbf{Gated}-UpConvResBlock & 32 & 3\\
\textbf{Gated}-ConvResBlock & 32 & 3\\
\hdashline
Conv2d & 32$\rightarrow$3 & 1 \\
Tanh() & 3 & 1 \\
\bottomrule
\end{tabular}
\label{table:resnet_g_scribble} 
\end{table}

\begin{table}[ht]
\caption{\textbf{Gated Resnet D:Scribble Dataset}}
\centering 
\begin{tabular}{l c c} 
\toprule
\textbf{Layer} & \textbf{Filter} & \textbf{Num Layers} \\
\midrule
Conv2d & 6 $\rightarrow$ 32 & 1\\
\hdashline
\textbf{Gated}-ConvResBlock & 32 & 3\\
\textbf{Gated}-DownConvResBlock & 32 & 4\\
\textbf{Gated}-ConvResBlock & 32 & 17\\
\hdashline
Conv2d & 32$\rightarrow$1 & 1 \\
Sigmoid() & 1 & 1 \\
\bottomrule
\end{tabular}
\label{table:resnet_d_scribble} 
\end{table}



\subsection{Gating Hypernetwork:}
The gating hypernetwork was also designed using Resnet blocks, we used 1D convolutions in the Resnet block \tabref{table:resblock1D} to reduce the number of parameters and to use BatchNormalization to speed up the training of the network responsible for the prediction of gating. The class conditioning is first passed through an embedding layer to obtain a representation of the class which could be further processed by the Resnet blocks. The same network is used for the various forms of gating. In case of block wise gating the number of outputs $dim^{gate}$ for this network is equal to the number of blocks used in the main network, in the case of an affine transformation the network predicts an equal number of biases for each of the block. In case of channel-wise gating the number of predicted parameters $dim^{gate}$ is equal to $num^{channels}\times num^{blocks}$ since each residual block consists of equal number of channels in each it helps to ease the training of the gating hypernetwork. The $\alpha$ was constrained between 0 and 1 to mimic whether to select a block or to reject the block while the $\beta$ in the case of affine transformation was restricted between -1 and 1. In the case of standard AdaIN, the parameters are unrestricted but it did not work in that setting and the AdaIN parameters had to be constrained between -1 and 1 in order for the network to perform well. 

\begin{table}[ht]
\caption{\textbf{ResBlock1D}} 
\centering 
\begin{tabular}{c} 
\toprule
\textbf{F(x)}\\\midrule
Conv1D\\ 
BatchNorm1D\\
ReLU \\
Conv1D\\
BatchNorm1D\\
ReLU \\
\bottomrule 
\end{tabular}
\label{table:resblock1D} 
\end{table}

\begin{figure*}[t]
    \centering
    \includegraphics[width=\linewidth,trim={0 0 4.5cm 0},clip]{paper_images/supplementary_grid_channel.pdf}
    \caption{{\bf Channel-Wise Gating:} We observe that the technique extends to not only the shapes it was trained on but can also generate images for some input shapes corresponding to other classes and to the extreme can generate images for certain shapes it never encountered during training such as the triangle and heart were directly downloaded from the internet. }
    \label{fig:channel_shapes}
    \vspace{-3mm}
\end{figure*}

\begin{table}[ht]
\caption{\textbf{Gating Hyper Network} $dim^{gate}$ is the number of blocks in the case of Block Wise Gating and the number of channels in the case of Channel Wise Gating. In case of affine its twice of each since the $\beta$ is of the same dimension }
\centering 
\begin{tabular}{l c c} 
\toprule
\textbf{Layer} & \textbf{Filter/Shape} & \textbf{Num Layers} \\
\midrule
Embedding & $dim^{embed}$ & 1 \\
Conv1d & 1 $\rightarrow$ 16 & 1\\
ResBlock1D & 16 & 16\\
Reshape & 16$\times dim^{embed}$ & 1\\
Linear & 16$\times dim^{embed} \rightarrow dim^{gate} $& 1\\
\bottomrule
\end{tabular}
\label{table:resnet_gating} 
\end{table}

\section{Unusual Shapes for Various Classes:}


As evident from \figref{fig:channel_shapes} 
the gated generative techniques extend to shapes it never was shown while training. 

\section{Subset of Comparison of Results:}

We show the results of our various gating mechanisms along with all the baselines reported in the paper on the Scribble Dataset's subset of test images. These images were the ones used for Amazon Mechanical Turk and the evaluation metrics reported in the paper.
\href{http://www.robots.ox.ac.uk/~arnabg/all_results_supplementary/index.html}{Comparison of our technique vs all baselines}

\section{Distribution of Alphas:}
\begin{figure*}[t]
    \centering
    \includegraphics[width=\linewidth]{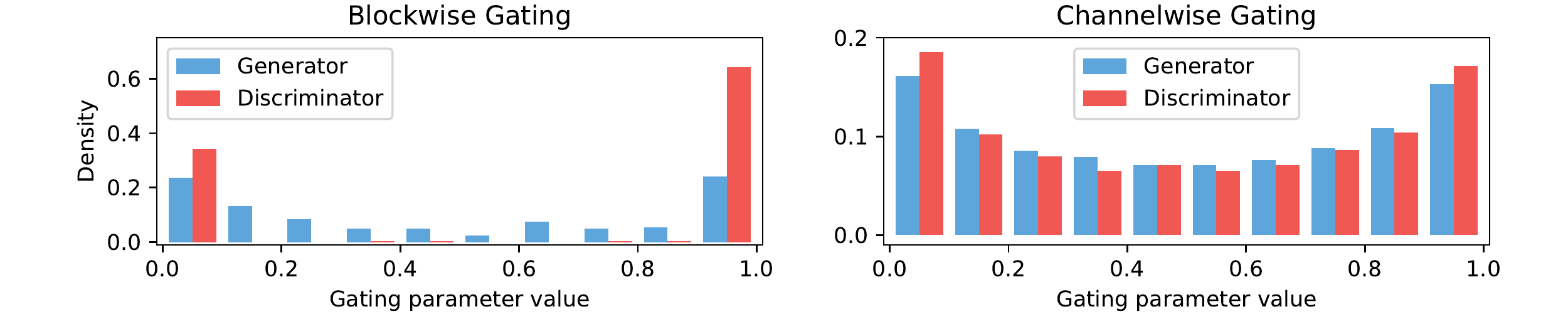}
    \caption{As we see from the distribution of the various alphas they are closer to the extremes (0 or 1) rather than the intermediate values, quite similar in the case of the channel wise gating as well }
    \label{fig:alpha_hist}
    \vspace{-3mm}
\end{figure*}

The histogram of the distribution of the various alphas for the block-wise setting and the channel-wise setting are shown in \figref{fig:alpha_hist}. Even without a sparsity constraint the alphas are pushed nearer the extremes rather than clustering near intermediate values. The effect is more starkly demonstrated in the case of the block wise gating, with the channel wise gating parameters being more evenly distributed.

\section{Interpolations:}
In order to judge the robustness of the trained models and to analyze the differences between the naive concatenation techniques compared with our gating mechanisms we conduct some inter-class interpolations. As we can see from \figref{fig:inter_watermelon_cookie}, \figref{fig:inter_orange_basketball}, \figref{fig:inter_cookie_moon} and \figref{fig:inter_orange_cupcake} that our gating mechanisms produce smooth transitions between classes while the case of naive concatenation techniques failing to generate basketball at all, thus failing at interpolating between basketball and orange.

{\small
\bibliographystyle{ieee}
\bibliography{src/gatedblocks}
}

\begin{figure*}[t]
    \centering
    \includegraphics[width=\linewidth]{interpolation-watermelon-cookie.pdf}
    \caption{{\bf Cookie $\rightarrow$ Watermelon:} As evident from the interpolation the gating produces much smoother transitions than simple concatenation techniques }
    \label{fig:inter_watermelon_cookie}
    \vspace{-3mm}
\end{figure*}

\begin{figure*}[t]
    \centering
    \includegraphics[width=\linewidth]{interpolation-orange-basketball.pdf}
    \caption{ {\bf Basketball $\rightarrow$ Orange:} A failure case of the simple conditioning technique, it never generates basketball and hence is not able to interpolate between basketball and orange. }
    \label{fig:inter_orange_basketball}
    \vspace{-3mm}
\end{figure*}

\begin{figure*}[t]
    \centering
    \includegraphics[width=\linewidth]{interpolation-cookie-moon.pdf}
    \caption{ {\bf Moon $\rightarrow$ Cookie:} Interpolation between moon and cookie }
    \label{fig:inter_cookie_moon}
    \vspace{-3mm}
\end{figure*}

\begin{figure*}[t]
    \centering
    \includegraphics[width=\linewidth]{interpolation-orange-cupcake.pdf}
    \caption{ {\bf Orange $\rightarrow$ Cupcake:} Interpolation between orange and cupcake, in the case of the baseline concat mechanism there's an abrupt transition from orange to cupcake while the transition is much smoother in the case of gated mechanisms }
    \label{fig:inter_orange_cupcake}
    \vspace{-3mm}
\end{figure*}

\begin{figure*}[t]
\centering
\begin{tabular}{*{2}{c@{\hspace{3px}}}}
\includegraphics[height=3.15cm,trim={6.0cm 0 7.6cm .4cm}, clip]{paper_images/alphas_chan_0.pdf} & 
\includegraphics[height=3.15cm,trim={5.8cm 0 .2cm .4cm}, clip]{paper_images/alpha_legend.pdf}
\\
\includegraphics[height=3.15cm,trim={6.0cm 0 7.6cm .4cm}, clip]{paper_images/alphas_chan_8.pdf} & 
\includegraphics[height=3.15cm,trim={5.8cm 0 .2cm .4cm}, clip]{paper_images/alpha_legend.pdf}
\\
\includegraphics[height=3.15cm,trim={6.0cm 0 7.6cm .4cm}, clip]{paper_images/alphas_chan_16.pdf} & 
\includegraphics[height=3.15cm,trim={5.8cm 0 .2cm .4cm}, clip]{paper_images/alpha_legend.pdf}
\\

\end{tabular}
\vspace{-2mm}
\caption{\label{fig:alpha_heat}
\textbf{Learned channel-wise gating parameters.} We show the soft-gating parameters for channelwise gating for the {\bf (top)} generator and {\bf (bot)} discriminator. Black indicates
completely off, and white indicates
completely on. We show all 24 blocks. This is an extension of Fig. 6 in the main paper. The nonuniformity of each columns indicates that different channels are used more heavily for different classes.
\vspace{-3mm}
}
\vspace{-2mm}
\end{figure*}